\documentclass[runningheads]{llncs}

\usepackage{graphicx}

\usepackage{tikz}
\usepackage{comment}
\usepackage{amsmath,amssymb} 
\usepackage{color}

\usepackage{subfigure}
\usepackage{url}
\usepackage{color}

\usepackage{booktabs}
\usepackage{amsfonts}
\usepackage{multirow}
\usepackage{soul}
\usepackage{wrapfig}
\usepackage{gensymb}
\usepackage{bm}
\usepackage{bbm}
\usepackage[utf8]{inputenc}
\usepackage[caption=false]{subfig} 
\newif\ifsubmit
\submitfalse

\ifsubmit
\newcommand{\xinyun}[1]{}
\newcommand{\aishan}[1]{}
\else
\newcommand{\xinyun}[1]{\textcolor{red}{[Xinyun: #1]}}
\newcommand{\aishan}[1]{\textcolor{blue}{[aishan: #1]}}
\fi

\begin{document}
\pagestyle{headings}
\mainmatter
\def\ECCVSubNumber{2686}  
\def\eg{\emph{e.g.}}
\def\ie{\emph{i.e.}}
\def\etc{\emph{etc}}
\def\etal{\emph{et al.}}
\def\wrt{\emph{w.r.t.}}

\title{Spatiotemporal Attacks for Embodied Agents} 


\titlerunning{Spatiotemporal Attacks for Embodied Agents}
%
\author{Aishan Liu\inst{1} \and
Tairan Huang\inst{1} \and
Xianglong Liu\inst{1,2}\protect\footnotemark[1] \and Yitao Xu\inst{1} \and Yuqing Ma\inst{1} \and Xinyun Chen\inst{3} \and Stephen J. Maybank \inst{4} \and Dacheng Tao\inst{5}}
\index{Maybank, Stephen J.}

%
\authorrunning{Liu et al.}
%
\institute{State Key Laboratory of Software Development Environment, \\Beihang University, China \and
Beijing Advanced Innovation Center for Big Data-Based Precision Medicine, \\
Beihang University, China
\and
UC Berkeley, USA
\and
Birkbeck, University of London, UK
\and
UBTECH Sydney AI Centre, School of Computer Science, Faculty of Engineering, \\
The University of Sydney, Australia
}
\maketitle

\renewcommand{\thefootnote}{\fnsymbol{footnote}}
\footnotetext[1]{Corresponding author. Email: xlliu@nlsde.buaa.edu.cn}

\begin{abstract}
Adversarial attacks are valuable for providing insights into the blind-spots of deep learning models and help improve their robustness. Existing work on adversarial attacks have mainly focused on static scenes; however, it remains unclear whether such attacks are effective against embodied agents, which could navigate and interact with a dynamic environment. In this work, we take the first step to study adversarial attacks for embodied agents. In particular, we generate spatiotemporal perturbations to form 3D adversarial examples, which exploit the interaction history in both the temporal and spatial dimensions. Regarding the temporal dimension, since agents make predictions based on historical observations, we develop a trajectory attention module to explore scene view contributions, which further help localize 3D objects appeared with highest stimuli. By conciliating with clues from the temporal dimension, along the spatial dimension, we adversarially perturb the physical properties (\eg, texture and 3D shape) of the contextual objects that appeared in the most important scene views. Extensive experiments on the EQA-v1 dataset for several embodied tasks in both the white-box and black-box settings have been conducted, which demonstrate that our perturbations have strong attack and generalization abilities.~\footnote{ Our code can be found at \url{https://github.com/liuaishan/SpatiotemporalAttack}.}

\end{abstract}

\section{Introduction}

Deep learning has demonstrated remarkable performance in a wide spectrum of areas \cite{krizhevsky2012imagenet,mohamed2011acoustic,sutskever2014sequence}, but it is vulnerable to adversarial examples \cite{szegedy2013intriguing,goodfellow6572explaining,chen2020boosting}. The small perturbations are imperceptible to human but easily misleading deep neural networks (DNNs), thereby bringing potential security threats to deep learning applications \cite{papernot2016practical,Liu2019Perceptual,Liu2020Biasbased}. Though challenging deep learning, adversarial examples are valuable for understanding the behaviors of DNNs, which could provide insights into the weakness and help improve the robustness \cite{zhang2019interpreting}. Over the last few years, significant efforts have been made to explore model robustness to the adversarial noises using \emph{adversarial attacks} in the static and non-interactive domain, \eg, 2D images \cite{goodfellow6572explaining,athalye2018obfuscated,Zhang2020PatchWise} or static 3D scenes \cite{zeng2019adversarial,liu2018beyond,xiao2019meshadv}. 

With great breakthroughs in multimodal techniques and virtual environments, embodied task has been introduced to further foster and measure the agent perceptual ability. An agent must intelligently navigate a simulated environment to achieve specific goals through egocentric vision \cite{das2018embodied,das2018neural,eqa_multitarget,gordon2018iqa}. For example, an agent is spawned in a random location within an environment to answer questions such as ``\emph{What is the color of the car?}''. Das \etal{} \cite{das2018embodied} first introduced the embodied question answering (EQA) problem and proposed a model consisting of a hierarchical navigation module and a question answering module. Concurrently, Gordon \etal{} \cite{gordon2018iqa} studied the EQA task in an interactive environment named AI2-THOR \cite{kolve2017ai2}. Recently, several studies have been proposed to improve agent performance using different frameworks \cite{das2018neural} and point cloud perception \cite{wijmans2019embodied}. Similar to EQA, embodied vision recognition (EVR) \cite{yang2019embodied} is an embodied task, in which an agent instantiated close to an occluded target object to perform visual object recognition.

\begin{figure}[!htb]
\vspace{-0.25in}
	\centering
	\includegraphics[width=0.7\linewidth]{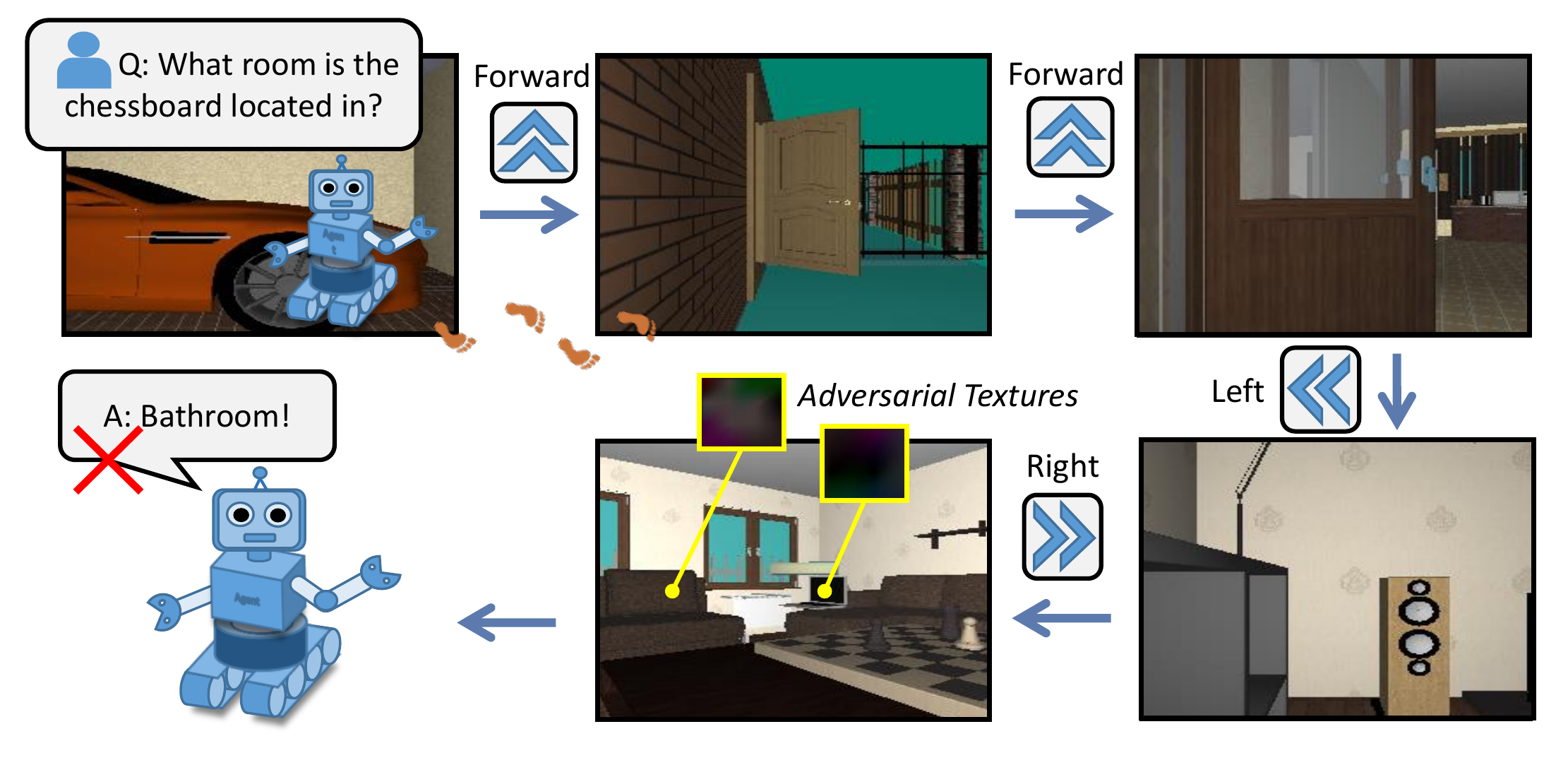}
	\caption{Embodied agents must navigate the environment through egocentric views to answer given questions. By adversarially perturbing the physical properties of 3D objects using our spatiotemporal perturbations, the agent gives the wrong answer (the correct answer is ``living room'') to the question. The contextual objects perturbed are: sofa and laptop.}
	\label{fig:startpage}
\vspace{-0.25in}
\end{figure}
In contrast to static tasks, embodied agents are free to move to different locations and interact with the dynamic environment. Rather than solely using a one-shot image, embodied agents observe 3D objects from different views and make predictions based on historical observations (trajectory). Current adversarial attacks mainly focused on the static scenes and ignored the information from the temporal dimension. However, since agents utilize contextual information to make decisions (\ie, answer questions), only considering a single image or an object appeared in one scene view may not be sufficient to generate strong adversarial attacks for the embodied agent.



In this work, we provide the first study of adversarial attacks for embodied agents in dynamic environments, as demonstrated in Figure \ref{fig:startpage}. By exploiting the interaction history in both the temporal and spatial dimensions, our adversarial attacks generate 3D spatiotemporal perturbations. Regarding the temporal dimension, since agents make predictions based on historical observations, we develop a trajectory attention module to explore scene view contributions, which could help to localize 3D objects that appeared with highest stimuli for agents' predictions. Coupled with clues from the temporal dimension, along the spatial dimension, we adversarially perturb the physical properties (\eg, 3D shape, and texture) of the contextual objects that appeared in the most important scene views. Currently, most embodied agents input 2D images transformed and processed from 3D scenes by undifferentiable renderers. To apply the attack using a gradient-based strategy, we replace the undifferentiable renderer with a differentiable one by introducing a neural renderer \cite{kato2018renderer}.

To evaluate the effectiveness of our spatiotemporal attacks, we conduct extensive experiments in both the white-box and black-box settings using different models. We first demonstrate that our generated 3D adversarial examples are able to attack the state-of-the-art embodied agent models and significantly outperform other 3D adversarial attack methods. Also, our adversarial perturbations can be transferred to attack the black-box renderer using non-differentiable operations, indicating the applicability of our attack strategy, and the potential of extending it to the physical world. We also provide a discussion of adversarial training using our generated attacks, and a perceptual study indicating that contrary to the human vision system, current embodied agents are mostly more sensitive to object textures rather than shapes,  which sheds some light on bridging the gap between human perception and embodied perception.

\section{Related Work}
\label{sec:work}

Adversarial examples or perturbations are intentionally designed inputs to mislead deep neural networks \cite{szegedy2013intriguing}. Most existing studies address the static scene including 2D images and static 3D scenes.

In the 2D image domain, Szegedy \emph{et al.} \cite{szegedy2013intriguing} first introduced adversarial examples and used the L-BFGS method to generate them. By leveraging the gradients of the target model, Goodfellow \emph{et al.} \cite{goodfellow6572explaining} proposed the Fast Gradient Sign Method (FGSM) which could generate adversarial examples quickly. In addition, Mopuri \emph{et al.} \cite{mopuri2018generalizable} proposed a novel approach to generate universal perturbations for DNNs for object recognition tasks. These methods add perturbations on 2D image pixels rather than 3D objects and fail to attack the embodied agents. 

Some recent work study adversarial attacks in the static 3D domain. A line of work \cite{xiao2019meshadv,zeng2019adversarial,liu2018beyond} used differentiable renderers to replace the undifferentiable one, and perform attacks through gradient-based strategies. They mainly manipulated object shapes and textures in 3D visual recognition tasks. On the other hand, Zhang \etal{} \cite{zhang2018camou} learned a camouflage pattern to hide vehicles from being detected by detectors using an approximation function. Adversarial patches \cite{brown2017adversarial,Liu2019Perceptual} have been studied to perform real-world 3D adversarial attacks. In particular, Liu \etal{} \cite{Liu2019Perceptual} proposed the PS-GAN framework to generate scrawl-like adversarial patches to fool autonomous-driving systems. However, all these attacks mainly considered the static scenes and ignored the temporal information. Our evaluation demonstrates that by incorporating both spatial and temporal information, our spatiotemporal attacks are more effective for embodied tasks. 

 Another line of work studies adversarial attacks against reinforcement learning agents \cite{Gleave2020AdversarialPA,Kos2017DelvingIA,Huang2017AdversarialAO,Pattanaik2018RobustDR,Lin2017TacticsOA}. These works mainly consider adversarial attacks against reinforcement learning models trained for standard game environments, where the model input only includes the visual observation. For example, most of existing work focuses on single-agent tasks such as Atari~\cite{bellemare2013arcade}, while Gleave \etal{}~\cite{Gleave2020AdversarialPA} studied adversarial attacks in multi-agent environments. Different from prior work, we focus on tasks related to embodied agents (i.e., EQA and EVR), with richer input features including both vision and language components. 

\section{Adversarial Attacks for the Embodiment}

The embodiment hypothesis is the idea that intelligence emerges in the interaction of an agent with an environment and as a result of sensorimotor activity \cite{smith2005development,das2018embodied}. To achieve specific goals, embodied agents are required to navigate and interact with the dynamic environment through egocentric vision. For example, in the EQA task, an agent is spawned at a random location in a 3D dynamic environment to answer given questions through navigation and interaction.

\subsection{Motivations}

Though showing promising results in the virtual environment, the agent robustness is challenged by the emergence of adversarial examples. Most of the agents are built
upon deep learning models which have been proved to be weak in the adversarial setting \cite{szegedy2013intriguing,goodfellow6572explaining}. By performing adversarial attacks to the embodiment, an adversary could manipulate the embodied agents and force them to execute unexpected actions. Obviously, it would pose potential security threats to agents in both the digital and physical world.

From another point of view, adversarial attacks for the embodiment are also beneficial to understand agents' behaviors. As black-box models, most deep-learning-based agents are difficult to interpret. Thus, adversarial attacks provide us with a new way to explore model weakness and blind-spots, which are valuable to understand their behaviors in the adversarial setting. Further, we can improve model robustness and build stronger agents against noises.


\subsection{Problem Definition}
In this paper, we use 3D adversarial perturbations (adversarial examples) to attack embodied agents in a dynamic environment.

In a \textbf{static scenario}, given a deep neural network $\mathbb F_{\theta}$ and an input image $\mathbf{I}$ with ground truth label $y$, an adversarial example $\mathbf{I}^{adv}$ is the input that makes the model conducted the wrong label
\begin{align*}
\mathbb F_{\theta}(\mathbf I^{adv}) \neq y  \quad s.t. \quad \|\mathbf I-\mathbf I^{adv}\| < \epsilon,
\end{align*}
where $\|\cdot\|$ is a distance metric to quantify the distance between the two inputs $\mathbf I$ and $\mathbf I^{adv}$ sufficiently small.


For the \textbf{embodiment}, an agent navigates the environment to fulfil goals and observe 3D objects in different time steps $t$. The input image $\mathbf I_t$ at time step $t$ for an agent is the rendered result of a 3D object from a renderer $\mathcal R$ by $\mathbf I_t = \mathcal R(\mathbf x,\mathbf c_t)$. $\mathbf x$ is the corresponding 3D object and $\mathbf c_t$ denotes conditions at $t$ (\eg, camera views, illumination, \etc.). To attack the embodiment, we need to consider the agent trajectory in temporal dimension and choose objects to perturb in the 3D spatial space. In other words, we generate adversarial 3D object $\mathbf x^{adv}$ by perturbing its physical properties at multiple time steps. The rendered image set $\{\mathbf I_{1},..., \mathbf I_N\}$ is able to fool the agent $\mathbb F_{\theta}$:
\begin{align*}
\mathbb F_{\theta} (\mathcal R(\mathbf x^{adv}_t,\mathbf c_t)) \neq y  \quad s.t. \quad \|\mathbf x_t- \mathbf x^{adv}_t\| < \epsilon,
\end{align*}
where $t$ belongs to a time step set we considered.

\begin{figure*}[!htb]
\vspace{-0.15in}
	\centering
	\includegraphics[width=1\linewidth]{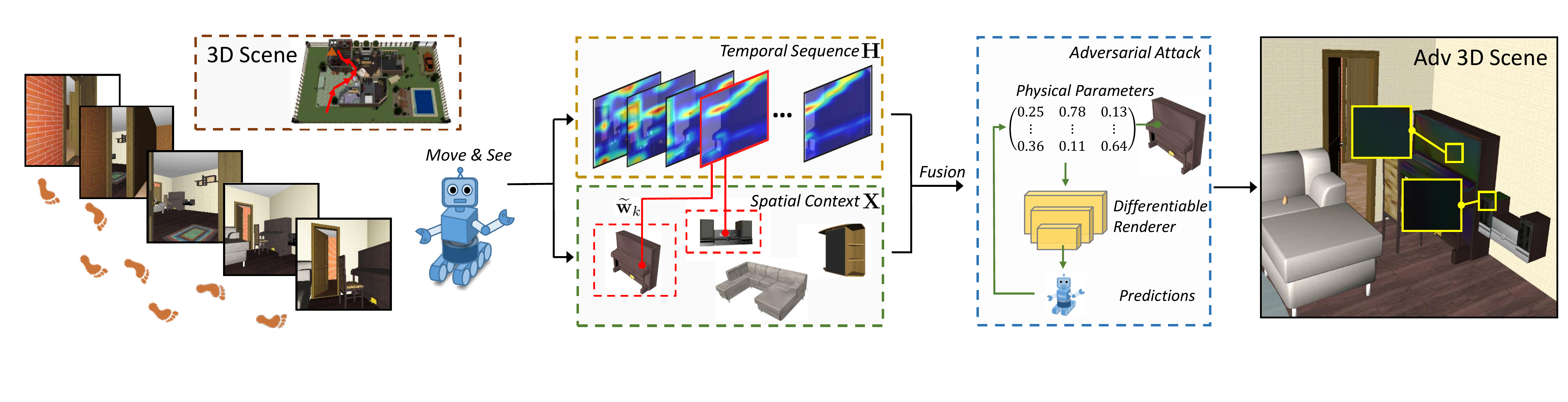}
	\caption{Our framework exploits interaction histories from both the temporal and the spatial dimension. In the temporal dimension, we develop a trajectory attention module to explore scene view contributions. Thus, important scene views are extracted to help localize 3D objects that appeared with highest stimuli for agents predictions. By conciliating with clues from the temporal dimension, along the spatial dimension, we adversarially perturb the 3D properties (\eg, 3D shape, and texture) of the contextual objects appeared in the most important scene views.}
	\label{fig:framework}
\vspace{-0.2in}
\end{figure*}

\section{Spatiotemporal Attack Framework}

In this section, we illustrate our framework to generate 3D adversarial perturbations for embodied agents in the dynamic environment. In Figure~\ref{fig:framework}, we present an overview of our attack approach, which incorporates history interactions from both the temporal and spatial dimensions.

Motivated by the fact that agents make predictions based on historical scene views (trajectory), we attack the 3D objects appeared in scene views containing the highest stimuli to the agent's prediction. In the temporal dimension, we develop a trajectory attention module  $\mathcal A$ to explore scene view contributions, which directly calculates the contribution weight for each time step scene view $\{\mathbf I_{1},..., \mathbf I_N\}$ to the agent prediction $\mathbb F_{\theta}$. Given a $N$-step trajectory, the $K$ most important historical scene views $\mathbf S$ are selected by $\mathcal A$ to help localize 3D objects that appeared with highest stimuli.

Meanwhile, rather than solely depending on single objects, humans always collect discriminative contextual information when making predictions. By conciliating with clues from the temporal dimension, along the spatial dimension, we adversarially perturb the physical properties $\bm{\mathbf{\phi}}$ of multiple 3D contextual objects $\mathbf X$ appeared in the most important scene views. Moreover, to attack physical properties (\ie, 3D shapes and textures), we also employ a differentiable renderer $\mathbb R_{\delta}$ to use the gradient-based attacks.

Thus, by coupling both temporal and spatial information, our framework generates spatiotemporal perturbations to form 3D adversarial examples, which could perform adversarial attacks for the embodiment. 



\subsection{Temporal Attention Stimulus}

To achieve specific goals, embodied agents are required to navigate the environment and make decisions based on the historical observations. Conventional vision tasks, \eg, classification, mainly base on one-shot observation in static images. In contrast, we should consider historical information (trajectory) such as last $N$ historical scene views observed by the agent $\mathbf H=\{\mathbf I_{t-N}, \mathbf I_{t-N+1},..., \mathbf I_{t-1}\}$, and adversarially perturb the 3D objects that appeared in them. Thus, we can formulate the attack loss:
\begin{equation}
\begin{aligned}
\mathcal L_{adv}(\mathbf H,y; \mathbb F_{\theta})= \mathrm P(y|\mathbf H),
\end{aligned}
\end{equation}

\noindent where $\mathrm P(\cdot|\cdot)$ denotes the prediction probability of the model, 
and $y$ indicates the ground truth label (\ie, correct answer, object class or action \wrt{} question answering, visual recognition and navigation, respectively). To attack agents, the equation above aims to decrease the confidence of the correct class.

There is extensive biological evidence that efficient perception requires both specialized visual sensing and a mechanism to prioritize stimuli, \ie, visual attention. Agents move their eyes towards a specific location or focus on relevant locations to make predictions by prioritizing different scene views \cite{carlone2018attention}. To improve attack abilities, we must design a visual attention module that selects a suitable set of visual features (historical scene views) to perform attack. Inspired by \cite{selvaraju2017grad}, given scene views $\mathbf H$, we first compute the gradient of target class $y$ \wrt{} normalized feature maps $\mathbf Z$ of a specified layer. These gradients flowing back are global average pooled to obtain weight $\mathbf{w}_{t}$ for the $t$-th scene view:
\begin{equation}
\begin{aligned}
\mathbf{w}_{t}= \text{max}(0, \sum^{r}_{n=1}\frac{1}{u \times v} \sum^{v}_{j=1} \sum^{u}_{i=1} \frac{\partial \mathrm P(y|\mathbf H)}{\partial \mathbf{Z}_{i,j}^{n}}),
\end{aligned}
\end{equation}
where $u \times v$ represents the size of the feature map, and $r$ indicates total feature map numbers in a specified layer. Then, We normalize each weight according to their mean vector $\bm{\mu}$ and variance vector $\bm{\sigma}$:
\begin{equation}
\begin{aligned}
\overline {\mathbf{w}}_{t} = \frac{\mathbf{w}_{t}-\bm{\mu}}{\bm{\sigma}^2+\epsilon},
\end{aligned}
\end{equation}

Thus, our trajectory attention module calculates the contribution of each scene view in the trajectory $\mathbf H$ towards the model decision for class $y$:
\begin{equation}
\begin{aligned}
\mathcal A(\mathbf H,y;\mathbb F_{\theta})=\langle \overline{\mathbf w}_{1},..., \overline{ \mathbf w}_{N}\rangle.
\end{aligned}
\end{equation}

The weights directly reflect the contribution of observed views at different time steps in the trajectory. Thus, we can further adversarially perturb the 3D objects that appeared in those scene views containing higher weights to execute a stronger attack.

\subsection{Spatially Contextual Perturbations}

Adversarial attacks in the static scene usually manipulate pixel values in the static image or different frames. In contrast, adversarial attacks for the embodiment require us to perturb the physical properties of 3D objects. Simply, we could randomly choose an object appeared in the most important scene views based on the attention weights to perform attacks. However, when humans look at an object, they always collect a discriminative context for that object \cite{garland2009starting}. In other words, we concentrate on that object while simultaneously being aware of its surroundings and context. The contextual information enables us to perform much stronger adversarial attacks. As shown in Figure \ref{fig:startpage}, when asking ``\emph{What room is the chessboard located in?}'', it is better to perturb contextual objects rather than only the target object ``chessboard''. To answer the question, agent relied on contextual objects (\eg, sofa, laptop, \etc), that convey critical factors and key features about the answer ``living room''.

Coupled with the clues from the temporal dimension, we further perturb the 3D contextual objects appeared in the $K$ most important views. Specifically, given $K$ most important scene views selected by our trajectory attention module $\mathbf S = \{\mathbf S_1,...,\mathbf S_K$\}, we perturb $M$ 3D objects $\mathbf X=\{\mathbf x_1, ...,\mathbf x_M\}$ appeared in $\mathbf S$. Thus, the adversarial attack loss can be formalized as:
\begin{equation}
\begin{aligned}
\mathcal L_{adv}(\mathbf X,y; \mathbb F_{\theta}, \mathbb{R_{\delta}})= \mathrm P(y|\mathbf S, \mathbb R_{\delta}(\mathbf X,\mathbf c)).
\end{aligned}
\end{equation}

Let $\bm {\phi_{m}}$ be the 3D physical parameters of object $\mathbf x_m$ (\eg, texture, shape, \etc{}). With the contribution weight $\overline{\mathbf w}$ for the $K$ most important scene views, we add the following perturbation to $\bm {\phi_{m}}$:
\begin{equation}
\begin{aligned}\small
\bm {\Delta\phi_{m}} = \sum_{k=1}^{K} \bm{\mathbbm{1}}(&\mathbf x_m \in \Phi(\mathbf S_k)) \cdot \overline{\mathbf w}_{k}\cdot \\
&\nabla_{\bm {\phi_{m}}} \mathcal L_{adv} (\mathbf x_m,y; \mathbb F_{\theta},\mathbb R_{\delta}), \\
\end{aligned}
\label{eq:physical-parameter-update}
\end{equation}
where $\Phi(\cdot)$ extracts the objects appeared in scene views.

\subsection{Optimization Formulations}

Based on the above discussion, we generate 3D adversarial perturbations using the optimization formulation:
\begin{equation}
\begin{aligned}\label{eqa:all}
\mathcal L (\mathbf X; \mathbb F_{\theta}, \mathbb R_{\delta}) =  \mathrm E_{\mathbf c \sim \mathbf C}& \bigg[ \mathcal L_{adv}(\mathbf X,y;\mathbb F_{\theta}, \mathbb R_{\delta},\mathbf c) + \\
& \lambda \cdot \mathcal L_{per}(\mathbf X, \mathbf X^{{adv}}; \mathbb R_{\delta}, \mathbf c)\bigg],
\end{aligned}
\end{equation}
where we append the adversarial attack loss with a perceptual loss:
\begin{equation}
\begin{aligned}
\mathcal L_{per}(\mathbf x, \mathbf x_{adv};\mathbb R_{\delta},\mathbf c)= ||\mathbb R_{\delta}(\mathbf x,\mathbf c)-\mathbb R_{\delta}(\mathbf x_{adv},\mathbf c)||,
\end{aligned}
\end{equation}
which constrains the magnitude of the total noises added to produce a visually imperceptible perturbation. $\mathbf C$ represents different conditions (\eg, camera views, illumination, \etc.) and $\lambda$ balances the contribution of each part.

Recent studies have highlighted that adversarial perturbations are ineffective to different transformations and environmental conditions (\eg, illuminations, rotations, \etc{}). In the dynamic environment, the viewing angles and environmental conditions change frequently. Thus, we further introduce the idea of \emph{expectation of transformations} \cite{athalye2017synthesizing} to enhance the attack success rate of our perturbations as shown in the expectation of different conditions $\mathbf C$ in Eqn (\ref{eqa:all}). Specifically, for each object to attack, we select five positional views one meter away with an azimuth angle uniformly ranging from [0\degree, 180\degree] to optimize the overall loss.

It is intuitive to directly place constraints on physical parameters such as the contour or color range of object surfaces. However, one potential disadvantage is that different physical parameters have different units and ranges. Therefore, we constrain the RGB intensity changes in the 2D image space after the rendering process to keep the consistency of the change of different parameters (\ie, shape or texture).


\section{Experiments}
In this section, we evaluate the effectiveness of our 3D spatiotemporal adversarial attacks against agents in different settings for different embodied tasks. We also provide a discussion of defense with adversarial training, and an ablation study of how different design choices affect the attack performance.

\subsection{Experimental Setting}
For both EQA and EVR tasks, we use the EQA-v1 dataset \cite{das2018embodied}, a visual question answering dataset grounded in the simulated environment. It contains 648 environments with 7,190 questions for training, 68 environments with 862 questions for validation, and 58 environments with 933 questions for testing. It divides the task into $T_{-10}$, $T_{-30}$, $T_{-50}$ by steps from the starting point to the target. We restrict the adversarial perturbations to be bounded by 32-pixel values per frame of size $224 \times 224$, in terms of $\ell_\infty$ norm.


\subsection{Evaluation Metrics}
To measure agent performance, we use evaluation metrics as in \cite{das2018embodied,wijmans2019embodied,das2018neural}:

- top-1 accuracy: whether the agent's prediction matches ground truth ($\uparrow$ is better);

- $d_T$: the distance to the target object at navigation termination ($\downarrow$ is better);

- $d_{\Delta}$: change in distance to target from initial to the final position ($\uparrow$ is better);

- $d_{min}$:  the smallest distance to the target at any point in the episode ($\downarrow$ is better);

Note that the goal of adversarial attacks is compromising the performance of the embodied agents, \ie, making worse values of the evaluation metrics above.

\subsection{Implementation Details}
\label{sec:implementation-details}
We use the SGD optimizer for adversarial perturbation generation, with momentum 0.9, weight decay $10^{-4}$, and a maximum of 60 iterations. For the hyper-parameters of our framework, we set $\lambda$ to 1, $K$ to 3, and $M$ as the numbers of all contextual objects observed in these frames. For EQA, we generate adversarial perturbations using PACMAN-RL+Q~\cite{das2018embodied} as the target model (we use ``PACMAN'' for simplicity), and we use Embodied Mask R-CNN~\cite{yang2019embodied} as the target model for EVR. In our evaluation, we will demonstrate that the attacks generated against one model could transfer to different models.

For both EQA and EVR, unless otherwise specified, we generate adversarial perturbations on texture only, \ie, in Equation~\ref{eq:physical-parameter-update}, we only update the parameters corresponding to texture, because it is more suitable for future extension to physical attacks in the real 3D environment. In the supplementary material, we also provide a comparison of adversarial perturbations on shapes, where we demonstrate that with the same constraint of perturbation magnitude, texture attacks achieve a higher attack success rate.

\begin{table*}
\centering
\scriptsize
\caption{Quantitative evaluation of agent performance on EQA task using different models in clean and adversarial settings (ours, MeshAdv \cite{xiao2019meshadv} and Zeng \etal{} \cite{zeng2019adversarial}). Note that the goal of attacks is to achieve a worse performance. We observe that our spatiotemporal attacks outperform the static 3D attack algorithms, achieving higher $d_T$ and $d_{min}$ as well as lower $d_{\Delta}$ and accuracy.}
\begin{tabular}{clrrrrrrrrrrrr}
   &&\multicolumn{9}{c}{\textbf{Navigation}}&\multicolumn{3}{c}{\textbf{QA}} \\
\cmidrule(lr){3-11} \cmidrule(lr){12-14}
&&\multicolumn{3}{c}{\bm{$d_{T}$} ($\downarrow$ is better)}&\multicolumn{3}{c}{\bm{$d_{\Delta}$} ($\uparrow$ is better)} & \multicolumn{3}{c}{\bm{$d_{min}$} ($\downarrow$ is better) }& \multicolumn{3}{c}{\textbf{accuracy}  ($\uparrow$ is better)  }\\
&&$T_{-10}$&$T_{-30}$&$T_{-50}$&$T_{-10}$&$T_{-30}$&$T_{-50}$&$T_{-10}$&$T_{-30}$&$T_{-50}$&$T_{-10}$&$T_{-30}$&$T_{-50}$\\
\hline
 \multirow{5}*{\textbf{PACMAN}}
 &Clean   &1.05&2.43&3.82&0.10&0.45&1.86&0.26&0.97&1.99&50.23\%&44.19\%&39.94\%\\
 &MeshAdv
 &1.06&2.44&3.90&0.09&0.44&1.78&0.31&1.17&2.33&16.07\%&15.34\%&13.11\%\\
  &Zeng \etal{}
  &\textbf{1.07}&2.46&3.88&\textbf{0.08}&0.42&1.80&0.42&1.37&2.43&17.15\%&16.38\%&14.32\%\\
  &\textbf{Ours}
  &1.06&\textbf{3.19}&\textbf{5.58}&0.09&\textbf{-0.39}&\textbf{0.10}&\textbf{0.90}&\textbf{2.47}&\textbf{5.33}&\textbf{6.17\%}&\textbf{4.26\%}&\textbf{3.42\%}\\
   \hline
   \multirow{5}*{\textbf{NAV-GRU}}
 &Clean   &1.03&2.47&3.92&0.12&0.41&1.76&0.34&1.02&2.07&48.97\%&43.72\%&38.26\%\\
 &MeshAdv &1.07&2.50&3.92&0.08&0.38&1.76&0.38&1.28&2.48&17.22\%&17.01\%&14.25\%\\
  &Zeng \etal{}
  &1.09&2.47&3.87&0.06&0.41&1.81&0.36&1.38&2.51&17.14\%&16.56\%&15.11\%\\
  &\textbf{Ours}
  &\textbf{1.13}&\textbf{2.96}&\textbf{5.42}&\textbf{0.02}&\textbf{-0.08}&\textbf{0.26}&\textbf{0.96}&\textbf{2.58}&\textbf{4.98}&\textbf{8.41\%}&\textbf{6.23\%}&\textbf{5.15\%}\\
   \hline
      \multirow{5}*{\textbf{NAV-React}}
 &Clean         &\textbf{1.37}&2.75&4.17&\textbf{-0.22}&0.13&1.51&0.31&0.99&2.08&48.19\%&43.73\%&37.62\%\\
 &MeshAdv       &1.05&2.79&4.25&0.10&0.09&1.43&0.32&1.29&2.47&15.36\%&14.78\%&11.29\%\\
 &Zeng \etal{}  &1.10&2.79&4.21&0.05&0.09&1.47&0.36&1.59&2.32&15.21\%&14.13\%&13.29\%\\
  &\textbf{Ours}&1.22&\textbf{2.85}&\textbf{5.70}&-0.07&\textbf{0.03}&\textbf{-0.02}&\textbf{1.06}&\textbf{2.59}&\textbf{5.47}&\textbf{8.26\%}&\textbf{5.25\%}&\textbf{5.39\%}\\
   \hline
      \multirow{5}*{\textbf{VIS-VGG}}
 &Clean  &1.02&2.38&3.67&0.13&0.50&2.01&0.38&1.05&2.26&50.16\%&45.81\%&37.84\%\\
 &MeshAdv &1.06&2.41&3.67&0.09&0.47&2.01&0.40&1.11&2.52&16.69\%&15.24\%&15.21\%\\
  &Zeng \etal{} &1.06&2.43&3.70&0.09&0.45&1.98&0.44&1.41&2.44&15.13\%&14.84\%&14.21\%\\
   &\textbf{Ours}
 &\textbf{1.18}&\textbf{2.83}&\textbf{5.62}&\textbf{-0.03}&\textbf{0.05}&\textbf{0.06}&\textbf{1.04}&\textbf{2.01}&\textbf{5.12}&\textbf{6.33\%}&\textbf{4.84\%}&\textbf{4.29\%}\\
   \hline
 \end{tabular}
 \label{tab:attack_result}
 \end{table*}

\subsection{Attack via a Differentiable Renderer}
we first provide the quantitative and qualitative results of our 3D adversarial perturbations on EQA and EVR through our differentiable renderer. For EQA, besides PACMAN, we also evaluate the transferability of our attacks using the following models: (1) NAV-GRU, an agent using GRU instead of LSTM in navigation~\cite{wijmans2019embodied}; (2) NAV-React, an agent without memory and fails to use historical information~\cite{das2018embodied}; and (3) VIS-VGG, an agent using VGG to encode visual information~\cite{das2018neural}. For EVR, we evaluate the white-box attacks on Embodied Mask R-CNN. As most of the embodied tasks can be directly divided into navigation and problem-solving stages, \ie, question answering or visual recognition, we attack each of these stages. We compare our spatiotemporal attacks to MeshAdv \cite{xiao2019meshadv} and Zeng \etal{} \cite{zeng2019adversarial}, both of which are designed for the static 3D environment, and thus do not leverage the temporal information.

\begin{figure}[htb]
	\centering
\subfigure[Clean Scene]{
\begin{minipage}[b]{0.47\linewidth}
\includegraphics[width=1\linewidth]{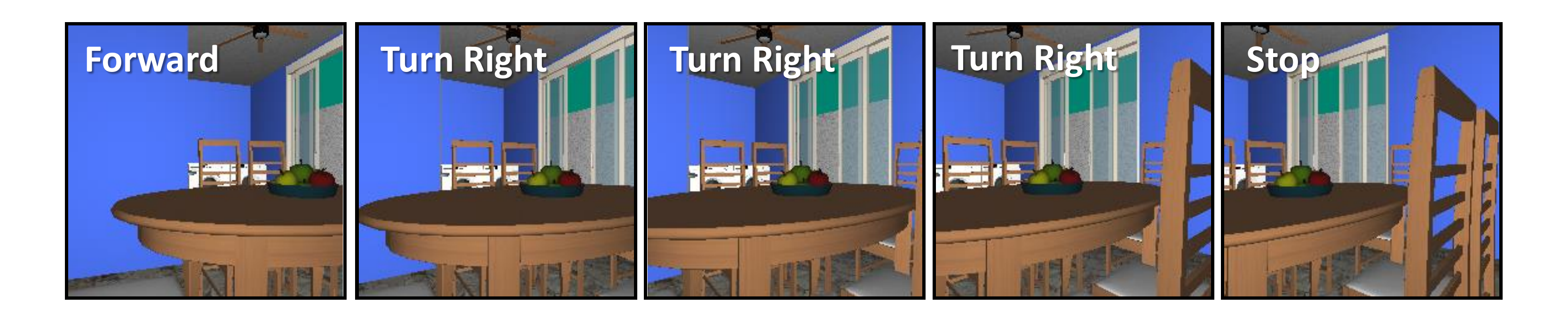}
\end{minipage}}
\subfigure[Adversarial Scene]{
\begin{minipage}[b]{0.47\linewidth}
\includegraphics[width=1\linewidth]{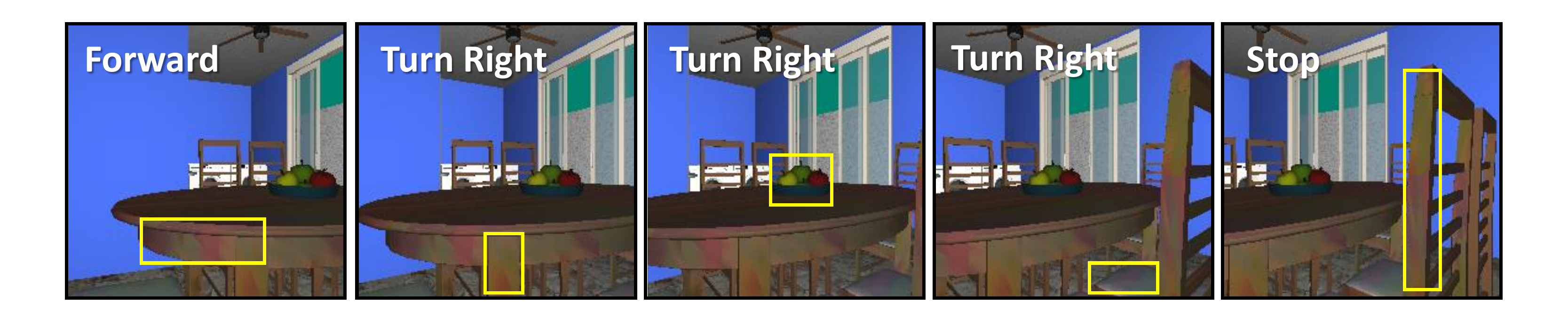}
\end{minipage}}
	\caption{Given the question ``\emph{What is next to the fruit bowl in the living room?}'', we show the last 5 views of the agent for EQA in the same scene with and without adversarial perturbations. The contextual objects perturbed including table, chairs and fruit bowel. The agent gives wrong answers ``television'' to the question (ground truth: chair) after seeing adversarial textures in subfigure (b). Yellow boxes show the perturbed texture.}
	\label{fig:perceptual}
\end{figure}

For \textbf{question answering} and \textbf{visual recognition}, we generate 3D adversarial perturbations using our proposed method on the test set and evaluate agent performance throughout the entire process, \ie, the agent is randomly placed and navigate to answer a question or recognize an object. As shown in Table \ref{tab:attack_result}, for white-box attacks, there is a significant drop in question answering accuracy from 50.23\%, 44.19\% and 39.94\% to 6.17\%, 4.26\% and 3.42\% for tasks with 10, 30, and 50 steps, respectively. Further, the visual recognition accuracy drastically decreases from 89.91\% to 18.32\%. The black-box attacks also result in a large drop in accuracy. The visualization of the last five steps before the agent's decision for EQA is shown in Figure \ref{fig:perceptual}. Our perturbations are unambiguous for human prediction but misleading to the agent.

For \textbf{navigation}, we generate 3D adversarial perturbations that intentionally stop the agent, \ie, make the agent predict \emph{Stop} during the navigation process. As shown in Table \ref{tab:attack_result}, for both white-box and black-box attacks, the values of $d_T$ and $d_{min}$ significantly increase compared to the clean environment when adding our perturbations, especially for long-distance tasks, \ie, $T_{-50}$. Further, the values of $d_\Delta$ decreases to around 0 after attack, which reveals that agents make a small number of movements to the destination. Also, some $d_\Delta$ even become negative, showing that the agent is moving away from the target.


%

To understand the transferability of attacks, we study attention similarities between models. The results can be found in the Supplementary Material.

In a word, our generated 3D adversarial perturbations achieve strong attack performance in both the white-box and black-box settings for navigation and problem-solving in the embodied environment.

\begin{figure}[!htb]
\centering
\subfigure[Accuracy]{
\begin{minipage}[b]{0.23\linewidth}
\includegraphics[width=1\linewidth]{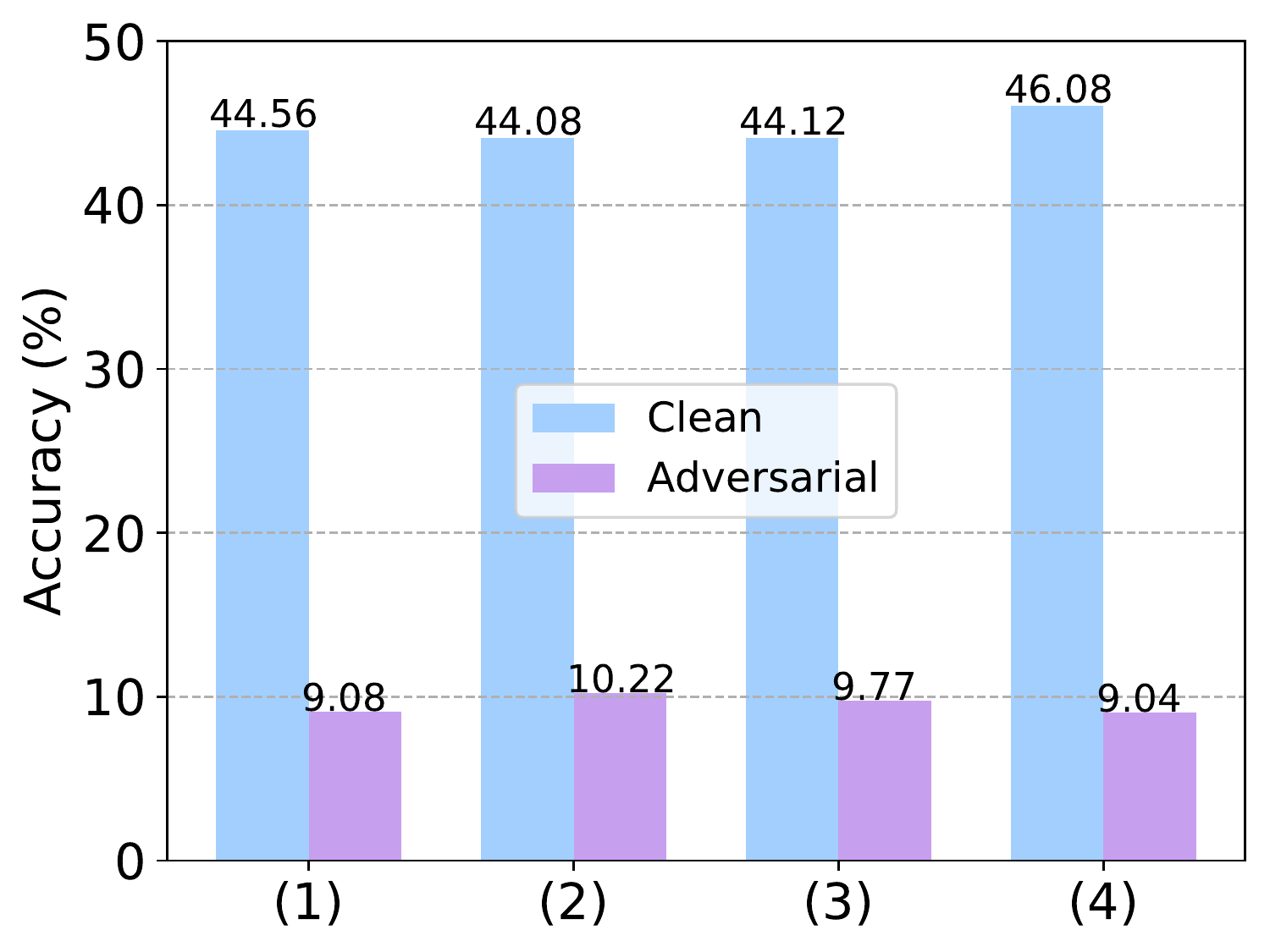}
\end{minipage}}
\subfigure[$d_T$]{
\begin{minipage}[b]{0.23\linewidth}
\includegraphics[width=1\linewidth]{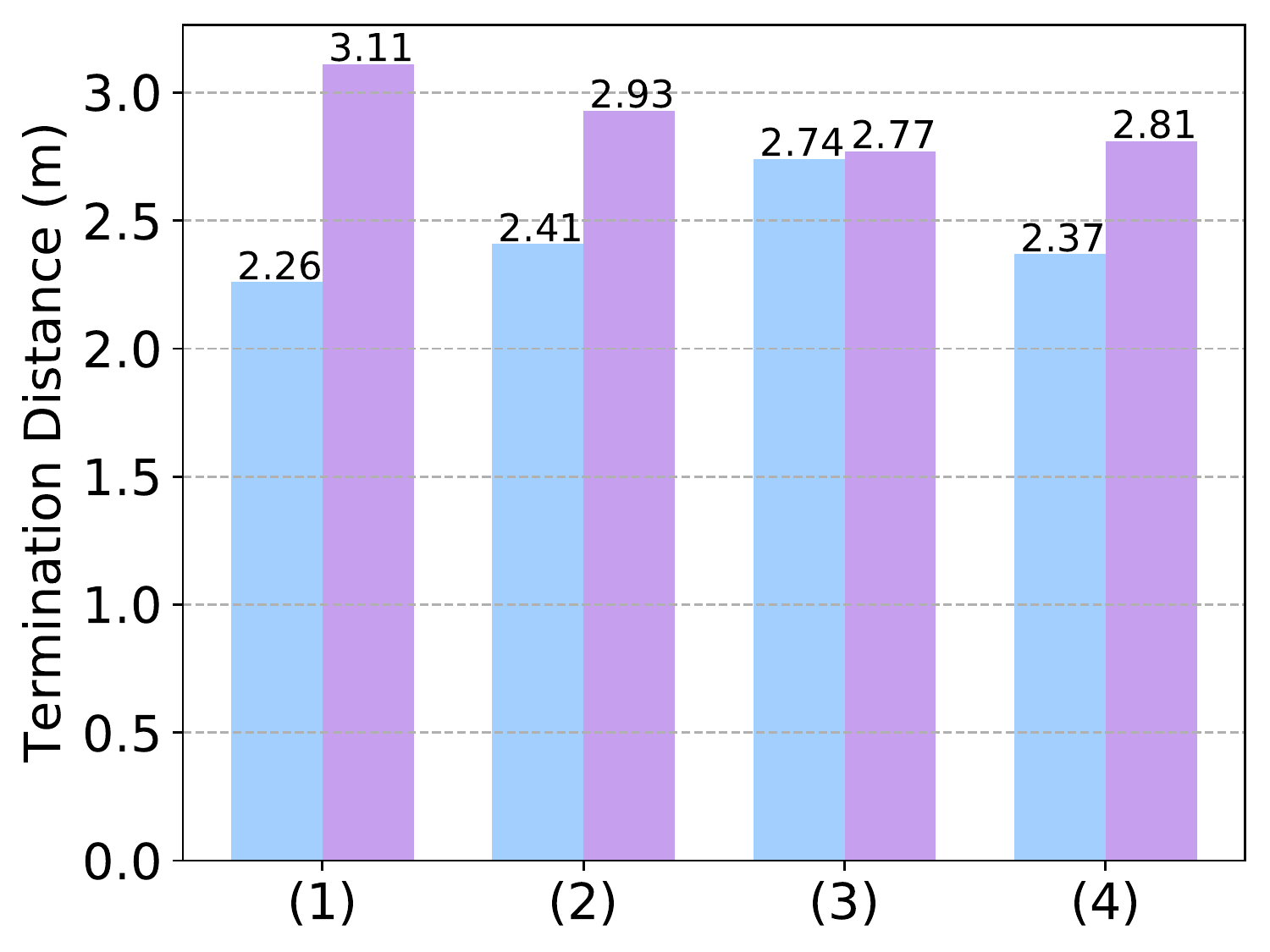}
\end{minipage}}
\subfigure[$d_{\Delta}$]{
\begin{minipage}[b]{0.23\linewidth}
\includegraphics[width=1\linewidth]{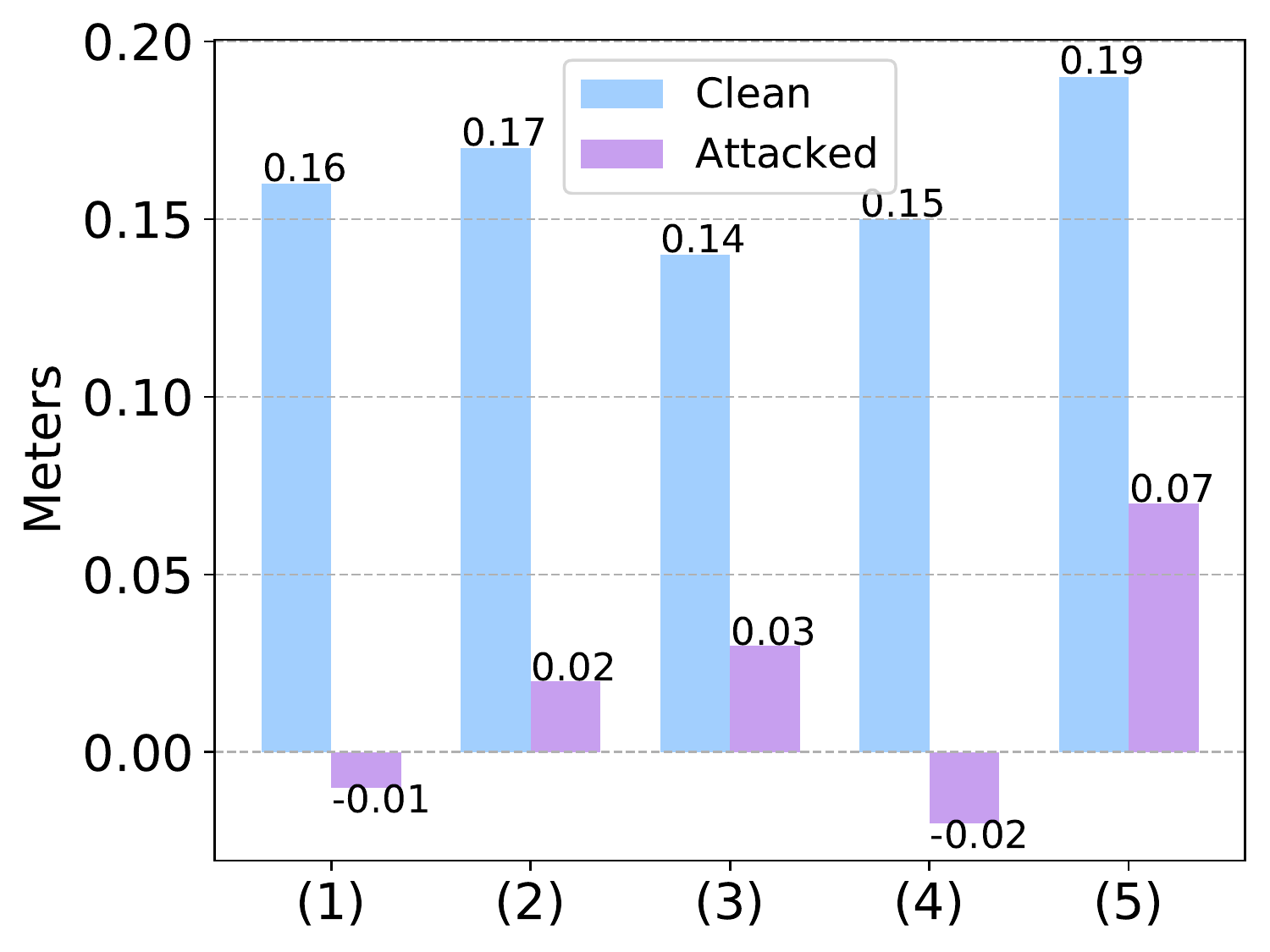}
\end{minipage}}
\subfigure[$d_{min}$]{
\begin{minipage}[b]{0.23\linewidth}
\includegraphics[width=1\linewidth]{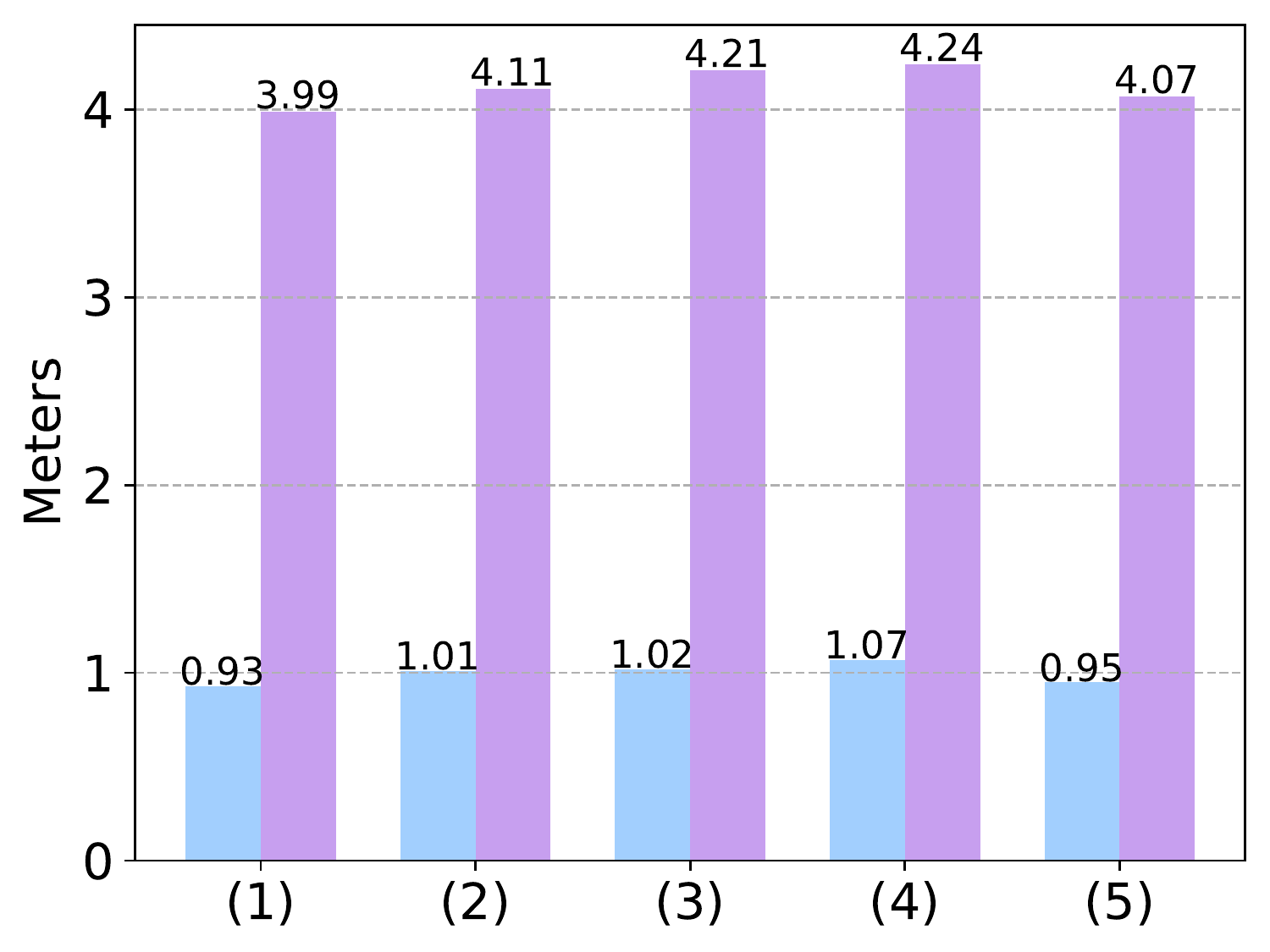}
\end{minipage}}

\caption{Method (1) to (4) represents PACMAN, NAV-GRU, NAV-React and VIS-VGG, respectively. Our framework generates adversarial perturbations with strong transferabilities to non-differentiable renderers.}
\label{fig:transfer}
\end{figure}
\subsection{Transfer Attack onto a non-differentiable Renderer}

Our proposed framework aims to adversarially attack $ \mathbb F_{\theta}(\mathbb R_{\delta} (\mathbf x_1, \mathbf x_2, ..., \mathbf x_n))$ by end-to-end gradient-based optimization. In this section, we further examine the potential of our framework in practice, where no assumptions about the non-differentiable renderer are given. By enabling interreflection and rich illumination, non-differentiable renderers can render images at high computational cost, so that the rendered 2D image is more likely to be an estimate of real-world physics. Thus, these experiments are effective to illustrate the transferability of generated adversarial perturbations and their potential in practical scenarios.

Specifically, we use the original non-differentiable renderer $\mathcal R$ for EQA-V1, which is implemented on OpenGL with unknown parameters, as the black-box renderer. We first generate 3D adversarial perturbations using our neural renderer $\mathbb R_{\delta}$, then save the perturbed scenes. We evaluate agent performance through the non-differentiable renderer $\mathcal R$ on those perturbed scenes to test the transferability of our adversarial perturbations.

As shown in Figure \ref{fig:transfer}, our spatiotemporal attacks can easily be transferred to a black-box renderer. However, our generated adversarial perturbations are less effective at attacking the non-differentiable renderer compared to the neural renderer. Many recent studies have reported that attacking the 3D space is much more difficult than attacking the image space \cite{zeng2019adversarial,xiao2019meshadv}. Further, we believe there are three other reasons for this phenomenon: (1) To generate attacks for the non-differentiable renderer, we first generate 3D adversarial perturbations using a differentiable renderer, then save the perturbed scenes into OBJ, MTL, and JPG files (the required files of the non-differentiable renderer to render a 3D scene) and feed them to the renderer. The information loss comes from the JPG compression process,  which may decrease the attack success rate. (2) The parameter difference between $\mathbb R_{\delta}$ and $\mathcal R$ may causes some minute rendering differences for the same scenarios. As adversarial examples are very sensitive to image transformations \cite{xie2017mitigating,guo2017countering}, the attacking ability is impaired; (3) The adversarial perturbation generated by optimization-based or gradient-based methods fails to obtain strong transferability due to either overfitting or underfitting \cite{dong2017boosting}.

\subsection{Generalization Ability of the Attack}
In this section, we further investigate the generalization ability of our generated adversarial perturbations. Given questions and trajectories, we first perturb the objects and save the scene. Then, loading the same perturbed scene, we ask agents different questions and change their start points to test their performance.

\begin{wraptable}{r}{4.6cm}
	\centering
	\scriptsize
	\begin{tabular}{lrrr}
	&\multicolumn{3}{c}{\textbf{QA accuracy}}
	\\
	& $T_{10}$ &	$T_{30}$ & $T_{50}$ \\
	\hline
	Clean & 51.42\% & 42.68\% & 39.15\% \\
	Attack& 6.05\%  & 3.98\%  & 3.52\%  \\
	Q & 10.17\%  & 8.13\%  & 7.98\%  \\
	T & 8.19\%  & 7.26\%  & 7.14\%  \\
	\hline
	\end{tabular}
	\caption{Generalization ability experiments. Our 3D perturbations generalize well in settings using different questions and starting points.}
	\label{table:generalization}
\end{wraptable}
 We first use the same perturbations on \textbf{different questions} (denoted as ``Q''). We fix the object in questions during perturbation generation and test to be the same. For example, we generate the perturbations based on question \emph{``What is the color of the table in the living-room?''} and test the success rate on question \emph{``What is next to the table in the living-room?''}. Moreover, we use the same perturbations to test agents from \textbf{different starting points} (\ie, different trajectories, denoted as ``T''). We first generate the perturbations and then test them by randomly spawning agents at different starting points (\ie, random rooms and locations) under the same questions. As shown in Table \ref{table:generalization}, the attacking ability drops a little compared to the baseline attack (generate perturbation and test at the scene with the same questions and starting point, denoted as ``Attack'') in both setting with higher QA accuracy but still very strong, which indicates the strong generalization ability of our spatiotemporal perturbations.

\subsection{Improving Agent Robustness with Adversarial Training}

 Given the vulnerability of existing embodied agents with the presence of adversarial attacks, we study defense strategies to improve the agent robustness. In particular, we base our defense on adversarial training~\cite{goodfellow6572explaining,Tu2019theoretical,madry2017towards}, where we integrate our generated adversarial examples for model training.

\textbf{Training.} We train 2 PACMAN models augmented with adversarial examples (\ie, we generate 3D adversarial perturbations on object textures, denoted as $AT$) or Gaussian noises (denoted as $GT$), respectively. We apply the common adversarial training strategy that adds a fixed number of adversarial examples in each epoch \cite{goodfellow6572explaining,alexey2017adversarialmachine}, and we defer more details in the supplementary material.

\begin{wraptable}{r}{4.6cm}
	\centering
	\scriptsize
	\begin{tabular}{lrrrr}
	&\multicolumn{2}{c}{\textbf{QA}} & \multicolumn{2}{c}{\textbf{Navigation}}\\
	\cmidrule(lr){2-3} \cmidrule(lr){4-5}
			&Adv & Gaussian& Adv & Gaussian  \\
	\hline
	Vanilla & 5.67\% & 22.14\% & 1.39 & 1.20 \\
	GT & 8.49\%  & 32.90\%  & 1.32 & 1.09\\
	\textbf{AT} & \textbf{23.56\%}  & \textbf{38.87\%}  & \textbf{1.17} & \textbf{1.01} \\
	\hline
	\end{tabular}
	\caption{Agent robustness in scenes with different noises. Adversarial training provides the most robust agent.}
	\label{table:advtrain}
\end{wraptable}
\textbf{Testing.} We create a test set of 110 questions in 5 houses. Following \cite{goodfellow6572explaining,hendrycks2018benchmarking}, we add different common noises including adversarial perturbations and Gaussian noises. To conduct fair comparisons, adversarial perturbations are generated in the white-box setting (\eg, for our adversarially trained model, we generate adversarial perturbations against it). 
The results in Table \ref{table:advtrain} support the fact that training on our adversarial perturbations can improve the agent robustness towards some types of noises (\ie, higher QA accuracy, and lower $d_T$).

\subsection{Ablation Study}
Next, we present a set of ablation studies to further demonstrate the effectiveness of our proposed strategy through different hyper-parameters $K$ and $M$, \ie, different numbers of historical scene views and contextual objects considered. All experiments in this section are conducted on $T_{-30}$.

\begin{wrapfigure}{r}{5.6cm}
\centering
\subfigure[]{
\begin{minipage}[b]{0.46\linewidth}
\includegraphics[width=1\linewidth]{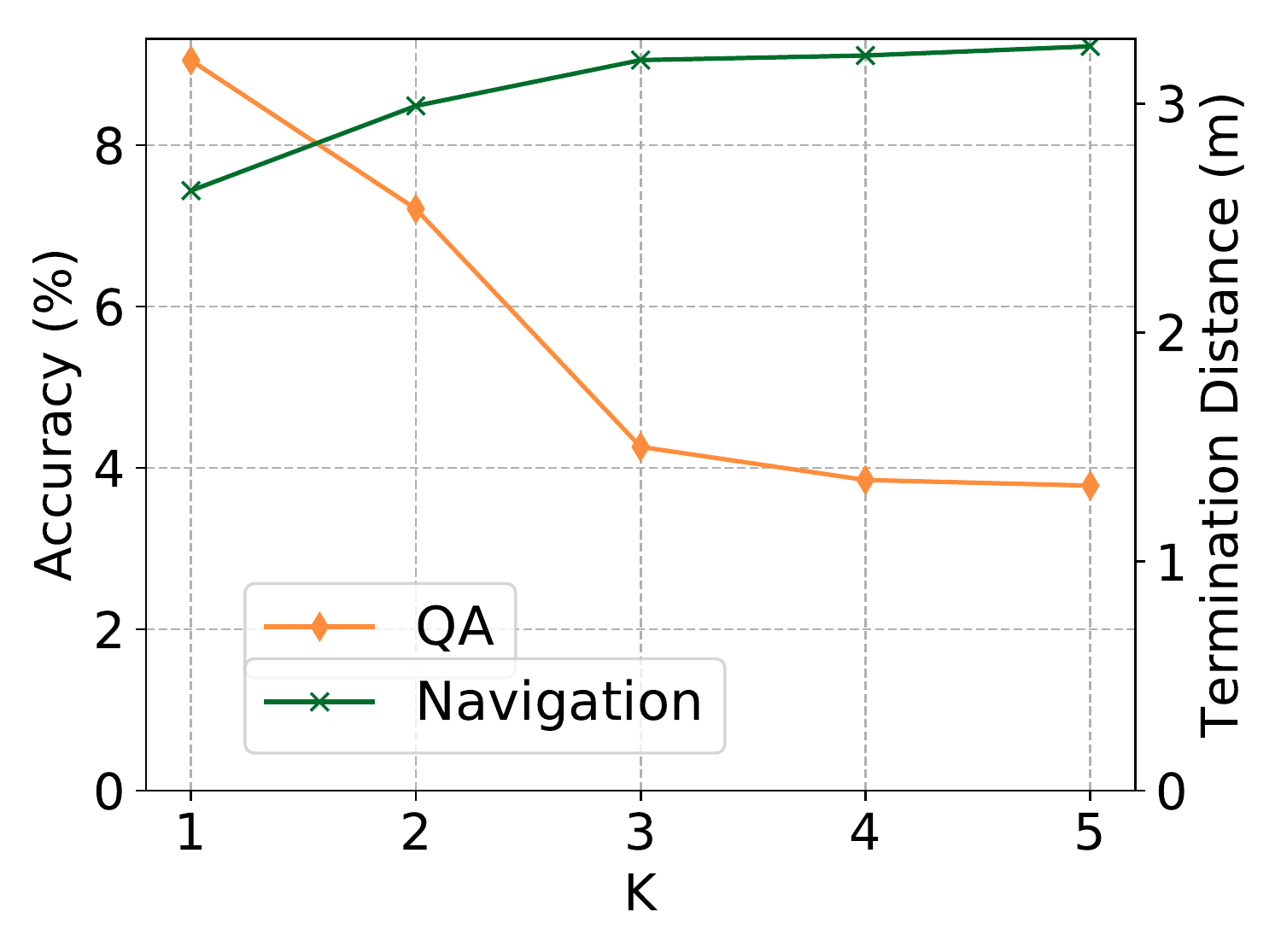}
\end{minipage}}
\subfigure[]{
\begin{minipage}[b]{0.46\linewidth}
\includegraphics[width=1\linewidth]{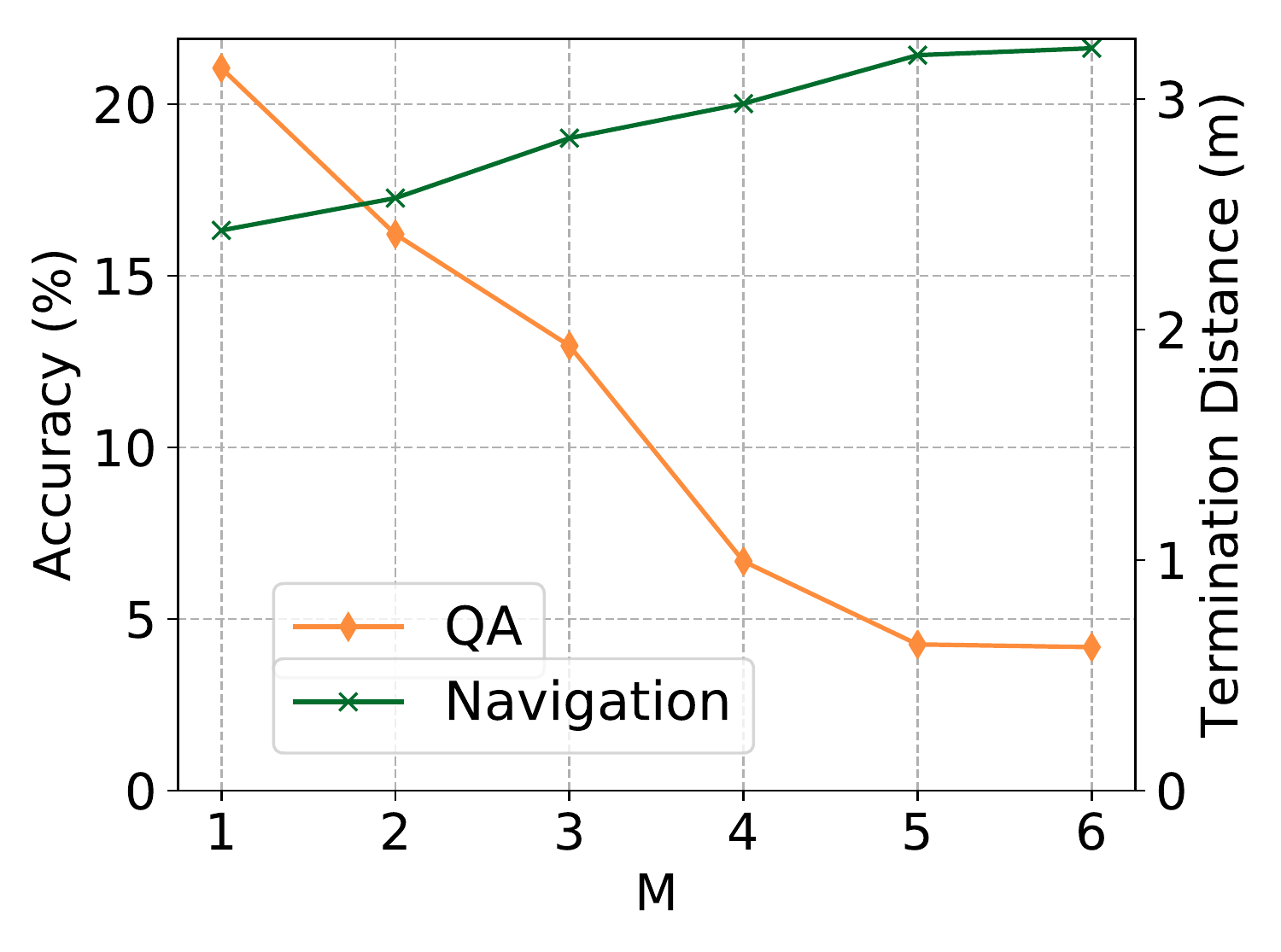}
\end{minipage}}
\caption{Ablation study with different $K$ and $M$ values in  (a) and (b).}
\label{fig:ablation}
\end{wrapfigure}

\textbf{Historical scene views numbers}. As for $K$, we set $K$=1,2,3,4,5, with a maximum value of $M$=5. For a fair comparison, we set the overall magnitude of perturbations to 32$/$255. As shown in Figure \ref{fig:ablation} (a), for navigation, we nearly obtain the optimal attack success rate when $K$=3. The results are similar to the question answering. However, the attack ability does not increase as significantly as that for navigation when increasing $K$. Obviously, the agent mainly depends on the target object and contextual objects to answer the questions. The contextual objects to be perturbed are quite similar to the increasing number of historical scene views considered.

\begin{figure}[!htb]
\centering
\subfigure[]{
\begin{minipage}[b]{0.46\linewidth}
\includegraphics[width=1\linewidth]{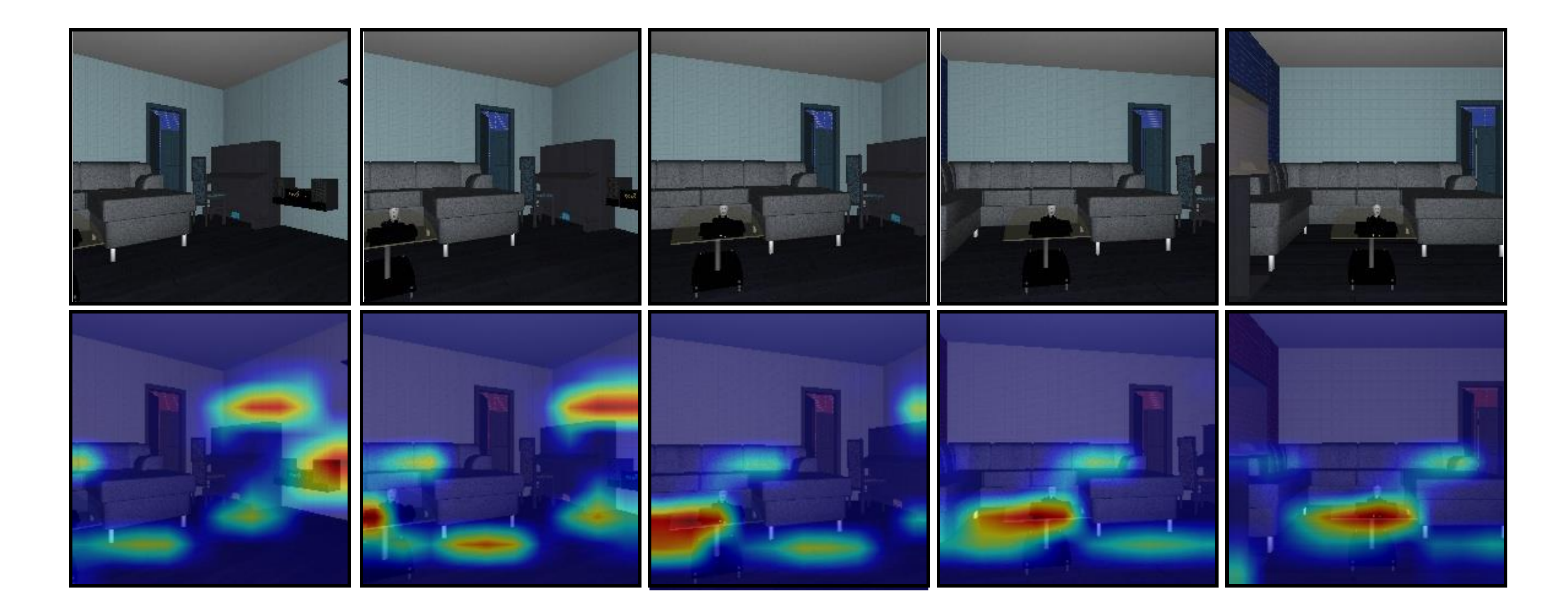}
\end{minipage}}
\subfigure[]{
\begin{minipage}[b]{0.46\linewidth}
\includegraphics[width=1\linewidth]{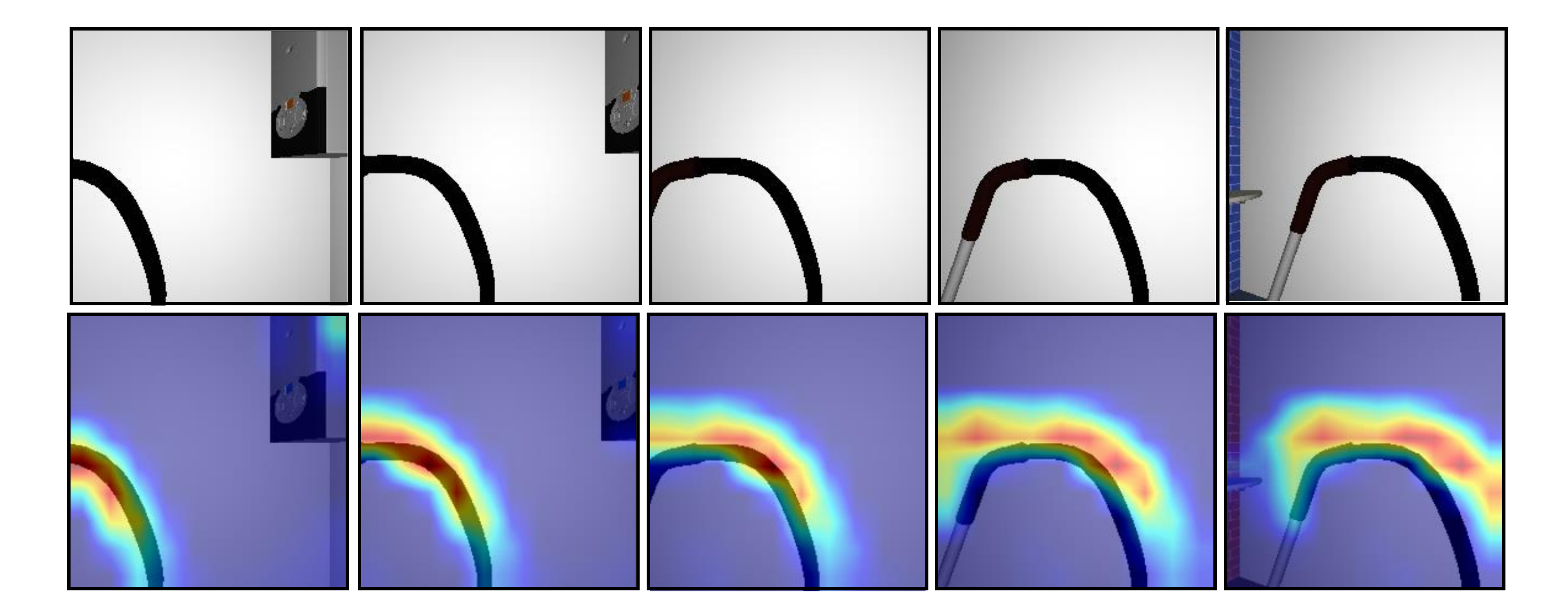}
\end{minipage}}
\caption{Visualization of last 5 views and corresponding attention maps. (a) denotes the locational and compositional question, and (b) denotes the color-related question.}
\label{fig:attention_vis}
\end{figure}

\textbf{Contextual objects numbers}. As for $M$, we set $M$=1,2,3,4,5,6 and $K$=3 to evaluate the contribution of the context to adversarial attacks. Similarly, we set the overall magnitude of adversarial perturbations to 32$/$255 for adversarial attacks with different $M$ values, \ie, perturbations are added onto a single object or distributed to several contextual objects. As shown in Figure \ref{fig:ablation}(b), the attack success rate increases significantly with the increasing of $M$ and converges at around 5. The reason is the maximum number of objects observable in 3 frames is around 5 or 6. Further, by considering the type of questions, we could obtain a deeper understanding about how an agent makes predictions. For questions about location and composition, \eg, ``\emph{What room is the $<$OBJ$>$ located in?}'' and ``\emph{What is on the $<$OBJ$>$ in the $<$ROOM$>$ ?}'', the attack success rate using context outperforms single object attack significantly with 4.67\% and 28.51\%, respectively. However, attacks on color-related questions are only 3.56\% and 9.88\% after contextual attack and single object attack, respectively. Intuitively, agents rely on different information to solve different types of questions. According to the attention visualization study shown in Figure \ref{fig:attention_vis}, agents generally utilize clues from contextual objects to answer locational and compositional questions while mainly focus on target objects when predicting their colors.


\section{Conclusion}
In this paper, we generate spatiotemporal perturbations to form 3D adversarial examples, which could attack the embodiment. Regarding the temporal dimension, we develop a trajectory attention module to explore scene view contributions, which further help localize 3D objects appeared with highest stimuli. By conciliating with clues from the temporal dimension, along the spatial dimension, we adversarially perturb the physical properties (\eg, texture) of the contextual objects that appeared in the most important scene views. Extensive experiments on the EQA-v1 dataset for several embodied tasks in both the white-box and black-box settings are conducted, which demonstrate that our framework has strong attack and generalization abilities. 

Currently, most embodied tasks could only be evaluated in the simulated environment. In the future, we are interested in performing spatiotemporal attacks in real-world scenarios. Using projection or 3D printing, we could bring our perturbations into the real-world to attack a real agent.

\section*{Acknowledgement}
This work was supported by National Natural Science Foundation of China (61872021, 61690202), Beijing Nova Program of Science and Technology\\ (Z191100001119050), Fundamental Research Funds for Central Universities (YWF-20-BJ-J-646), and ARC FL-170100117.

\clearpage
%
%
\bibliographystyle{splncs04}
\bibliography{egbib}

\pagestyle{headings}
\mainmatter
\def\ECCVSubNumber{2686}  
\def\eg{\emph{e.g.}}
\def\ie{\emph{i.e.}}
\def\etc{\emph{etc}}
\def\etal{\emph{et al.}}
\def\wrt{\emph{w.r.t.}}

\title{Supplementary Material: Spatiotemporal Attacks for Embodied Agents} 

\author{Aishan Liu\inst{1} \and
Tairan Huang\inst{1} \and
Xianglong Liu\inst{1,2}\protect\footnotemark[1] \and Yitao Xu\inst{1} \and Yuqing Ma\inst{1} \and Xinyun Chen\inst{3} \and Stephen J. Maybank \inst{4} \and Dacheng Tao\inst{5}}
\index{Maybank, Stephen J.}

%
\authorrunning{Liu et al.}
%
\institute{State Key Laboratory of Software Development Environment, \\Beihang University, China \and
Beijing Advanced Innovation Center for Big Data-Based Precision Medicine, \\
Beihang University, China
\and
UC Berkeley, USA
\and
Birkbeck, University of London, UK
\and
UBTECH Sydney AI Centre, School of Computer Science, Faculty of Engineering, \\
The University of Sydney, Australia
}

\titlerunning{Spatiotemporal Attacks for Embodied Agents}
\maketitle

\section{More Details of Experimental Settings}

\subsection{Generalization Ability of the Attack}
We provide the details of our experimental settings in Section 5.6. We use 99 questions of 5 houses for both of the following 2 settings.

\textbf{Different questions.} For each question, we generate spatiotemporal perturbations based on the current question, and then evaluate another question given the scenes with the same perturbed object.

\textbf{Different starting points.} For each question, we randomly sample another question, and then use its starting point as the initilization point for the agent to answer the current question. The average distance change for the starting point is 2.47 (maximum is 9.52, minimum is 0.83). Among the 99 new starting points, 45.45\% of them are in the different rooms.

\subsection{Improving Agent Robustness with Adversarial Training}

We provide the details of our experimental settings in Section 5.7, where we evaluate the effectiveness of adversarial training for $T_{-10}$ setting.

\textbf{Training}. We use the SGD optimizer for adversarial training, with a learning rate of 0.001. Following \cite{das2018embodied}, both the QA and NAV modules are trained for 300 epochs. In each training batch, we generate one perturbed scene (either adding the adversarial perturbation or the Gaussian noise) for each clean scene, so that the numbers of clean scenes and perturbed scenes are the same per batch, \ie, 4 clean scenes and 4 perturbed scenes per batch in our experiments. The magnitude of adversarial perturbations is 32$/$255. For Gaussian noises, we choose the maximum noise severity level following \cite{hendrycks2018benchmarking}, and set the mean to be 0, the standard deviation to be 0.38. The other settings are the same as that in Section 5.3.


\textbf{Testing}. For evaluation, we use the same approaches as for training to add either adversarial perturbations or Gaussian noises to the chosen 3D objects.

\subsection{Perceptual Studies via Amazon Mechanical Turk (AMT)}

We design a perceptual study on AMT to figure out which features are more sensitive and attractive for human predictions, \ie, shape or texture. For each question, the participants need to select the correct category of the object in the picture. We do not set any time limit for the responses.

%

In total, we collect 30 objects in different scenes, each of which is perturbed on shape and texture, respectively. Thus, we have a total of 60 questions, namely 60 Human Intelligence Tasks (HITs). For each HIT, we make 10 assignments, \ie, each HIT will be answered by 10 different human workers. As a result, we finally collect 600 responses for our perceptual study.

For fair comparisons, we use the same hyper-parameters for shape and texture attacks. We limit the overall perturbation magnitude to 32/255, as in other settings.

\section{Additional Experimental Results}

In this section, we provide more results of our attacks on a black-box renderer, as well as more examples and analysis of our attacks.

\subsection{Texture v.s. Shape}
\label{sec:texture-shape}

In this section, we study the importance of texture and shape for model predictions. For a fair comparison, we set the same constraint of perturbation magnitude for both texture and shape attacks, as in Section~\ref{sec:implementation-details}. According to the accuracy of the texture attack (4.26\%) and shape attack (27.14\%) in the $T_{-30}$ task, perturbing textures is far more effective than perturbing shapes. A question emerges: \emph{Which is more important for model prediction, texture or shape?}

A recent study \cite{geirhos2018imagenet} demonstrated that CNNs are strongly biased towards recognizing textures. Compared to long-range dependencies encoded in the shapes of objects, standard CNNs prefer local textures \cite{zhang2019interpreting}. Thus, it is not uncommon to see that the agent is more likely to make errors when 3D object textures are adversarially perturbed.

\begin{wrapfigure}{r}{4.6cm}
\vspace{-0.15in}
	\centering
	\includegraphics[width=1\linewidth]{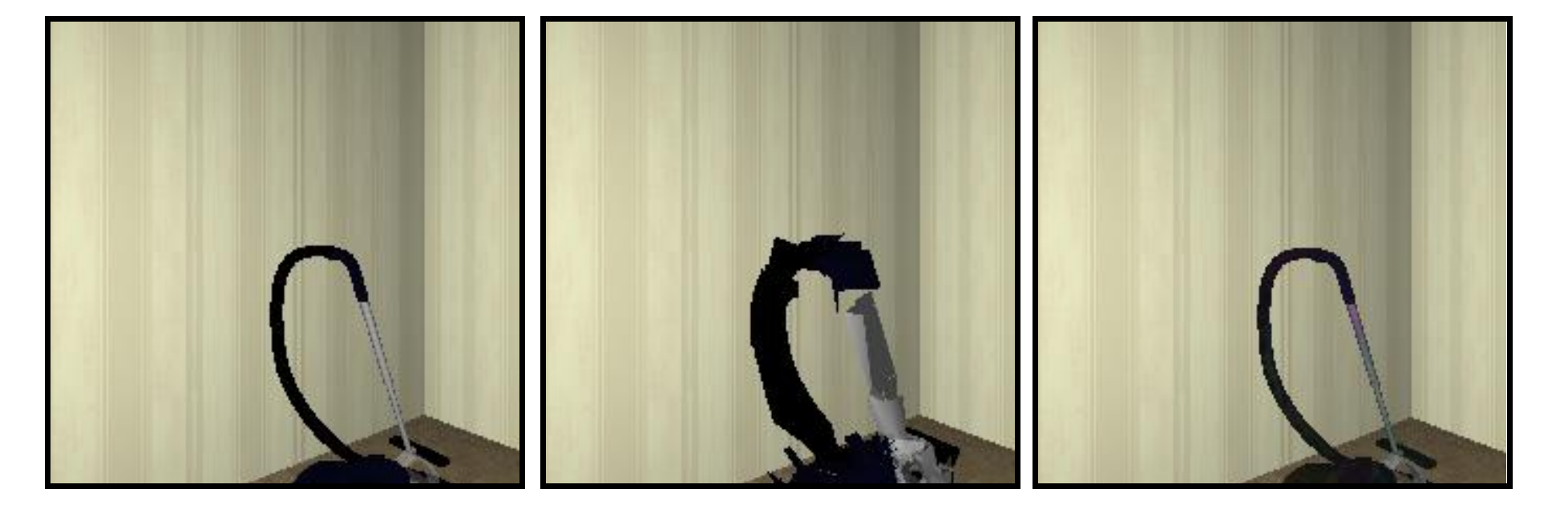}
	\caption{Visualization of scene perturbed on different physical parameters. From left to right: clean, shape attacks, and texture attacks.}
	\label{fig:amt}
\vspace{-0.15in}
\end{wrapfigure}

Since deep learning prefers textural information when making decisions, it is worth studying which features humans find more beneficial. As a preliminary step, we examined which features are more sensitive for human predictions with a user study conducted on the Amazon Mechanical Turk (AMT). With each object adversarially perturbed in texture and shape (see Figure \ref{fig:amt}), participants were asked to assign those adversarial objects to one of five classes (the ground-truth class, the top-3 adversarial target classes, and ``none of the above''). Our results showed that the classification accuracy for adversarial texture manipulation (83.3\%) was higher than that for shape (32.7\%). It indicates that shape is a more sensitive parameter for human predictions compared to texture. This is obvious since people are more likely to recognize a table with different textures rather than a table made out of wood but showing a strange shape.

In conclusion, embodied agents trained upon most current strategies are more sensitive to texture rather than shape. It is in stark contrast to humans and reveals fundamental differences in classification strategies between humans and machines. Therefore, to bridge the gap between human perception and embodied perception, it is important to train agents that can better capture shape-based features. Could we obtain stronger policies for agents if we train them with shape-based adversarial perturbations? We put it as future work.

\subsection{Transfer Attack onto a Black-box Renderer}

\begin{figure}[htbp]
\centering
\subfigure[Accuracy]{
\begin{minipage}[b]{0.23\linewidth}
\includegraphics[width=1\linewidth]{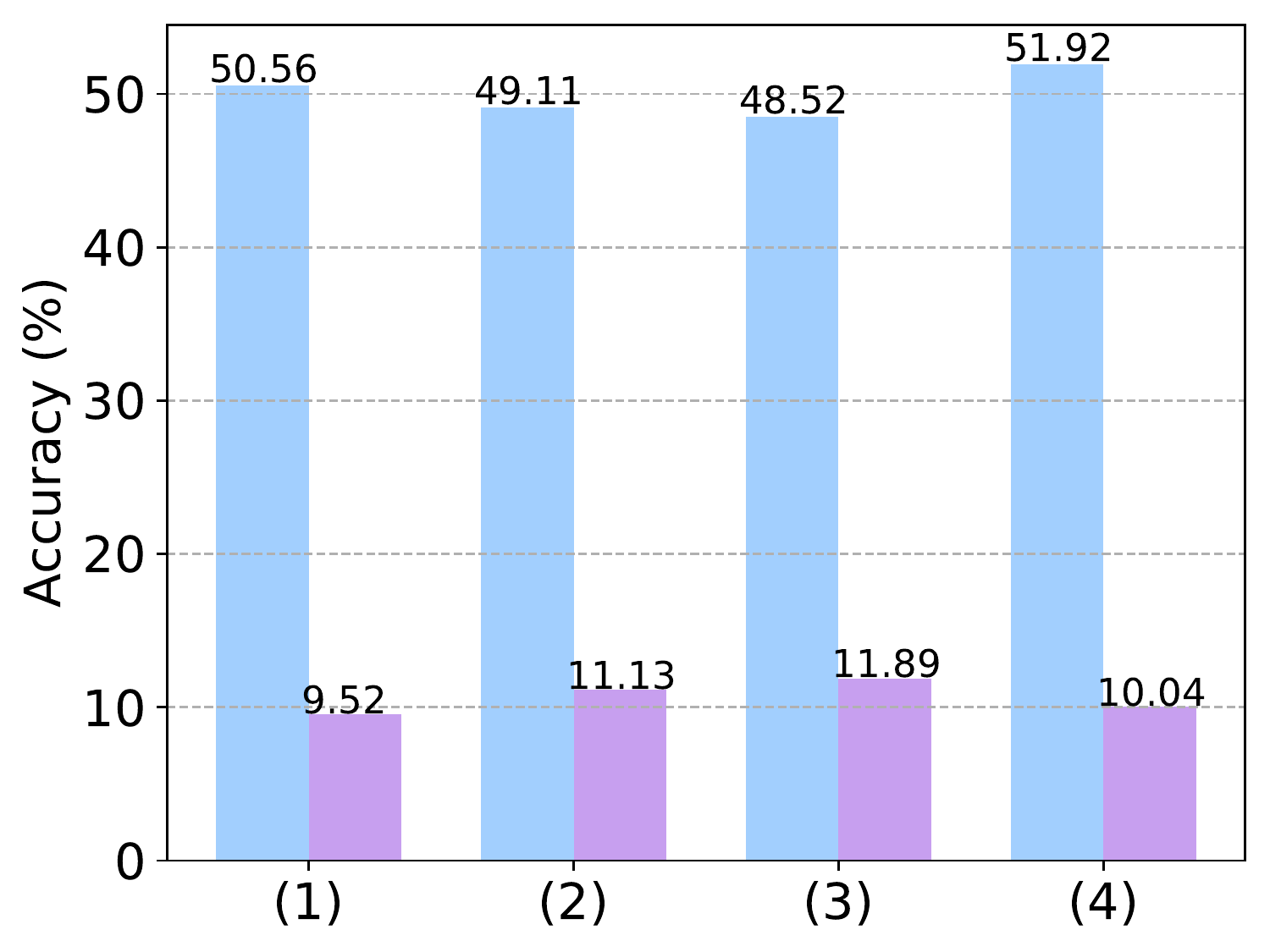}
\end{minipage}}
\subfigure[$d_T$]{
\begin{minipage}[b]{0.23\linewidth}
\includegraphics[width=1\linewidth]{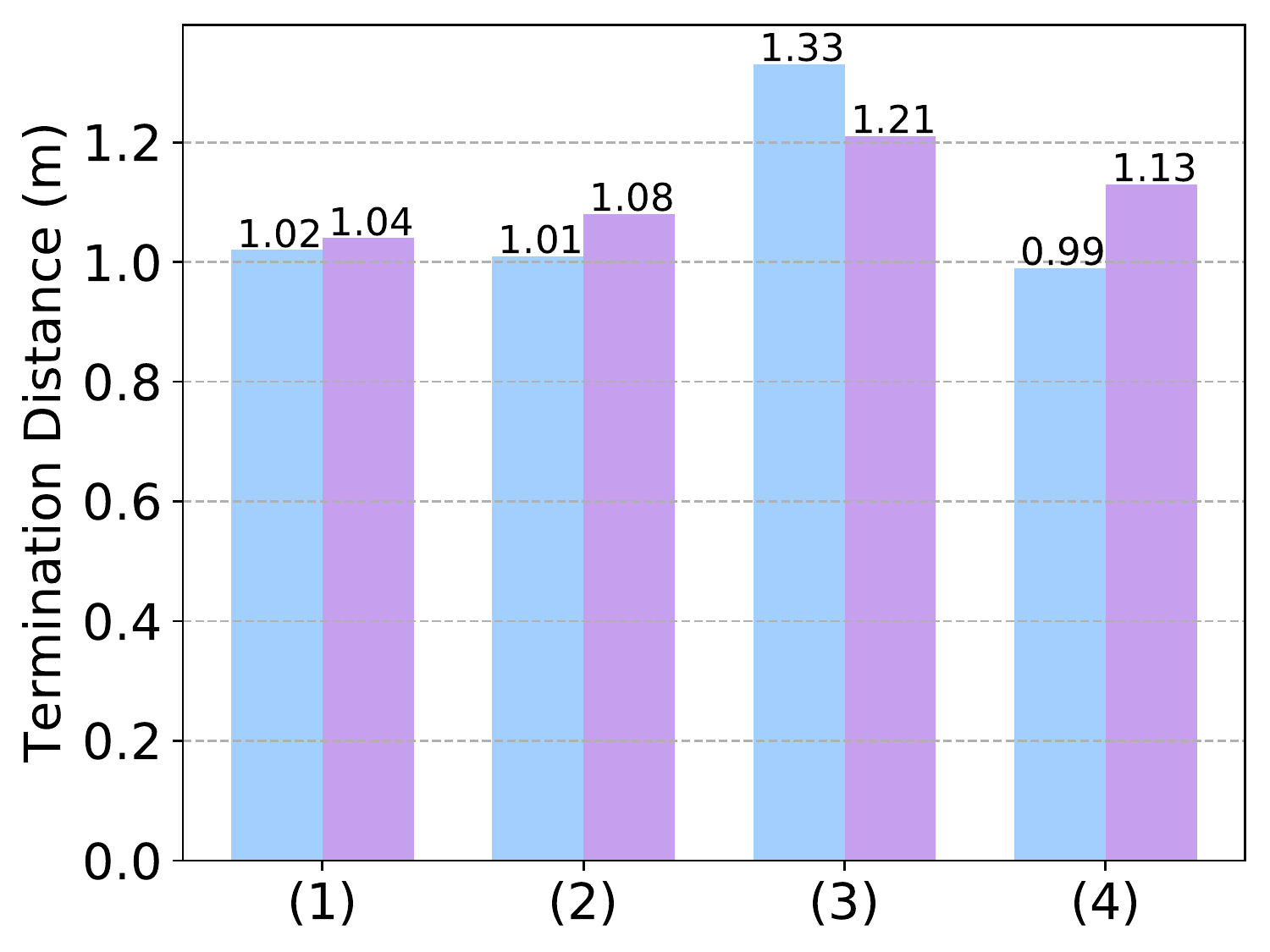}
\end{minipage}}
\subfigure[$d_{\Delta}$]{
\begin{minipage}[b]{0.23\linewidth}
\includegraphics[width=1\linewidth]{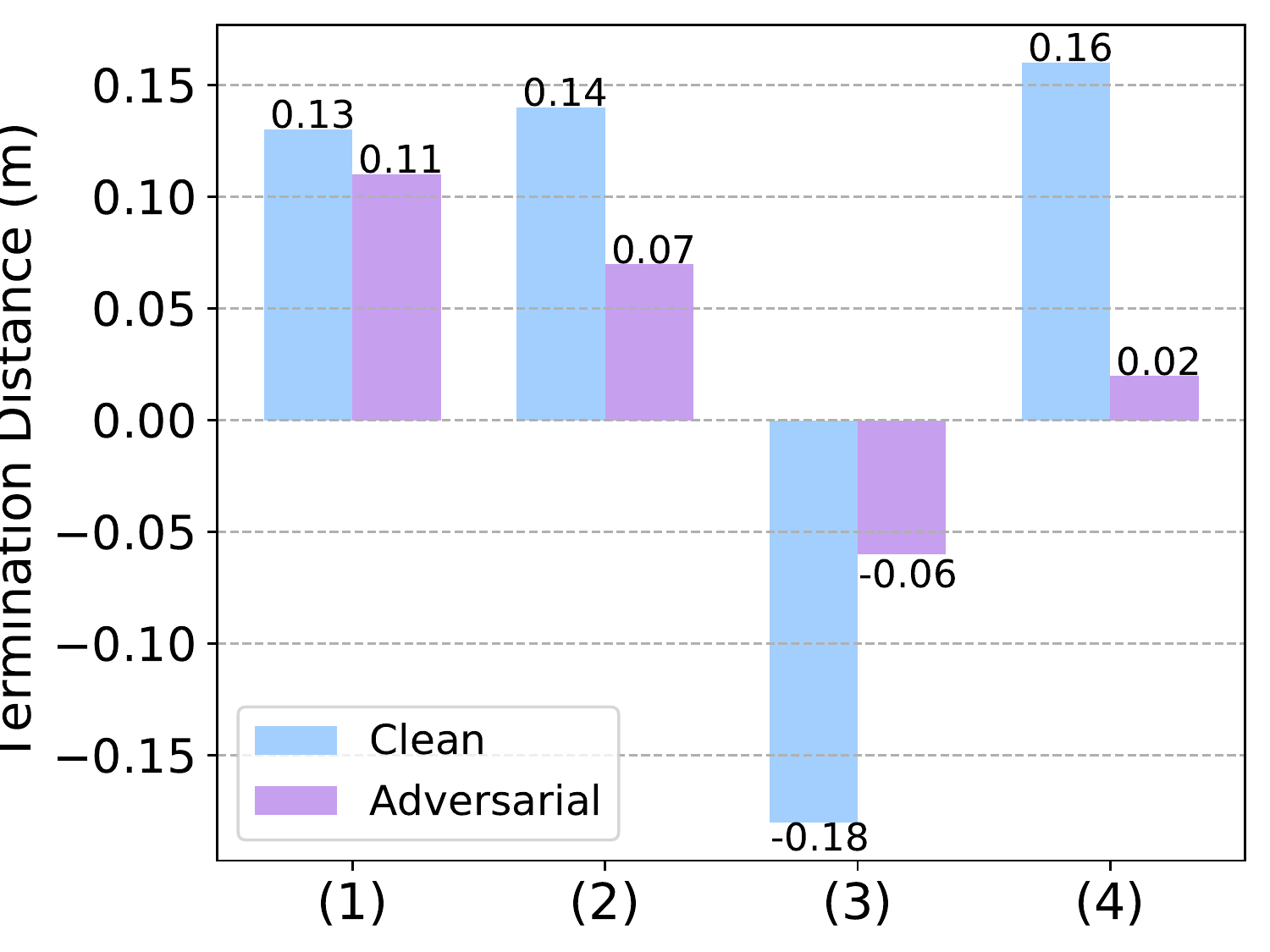}
\end{minipage}}
\subfigure[$d_{min}$]{
\begin{minipage}[b]{0.23\linewidth}
\includegraphics[width=1\linewidth]{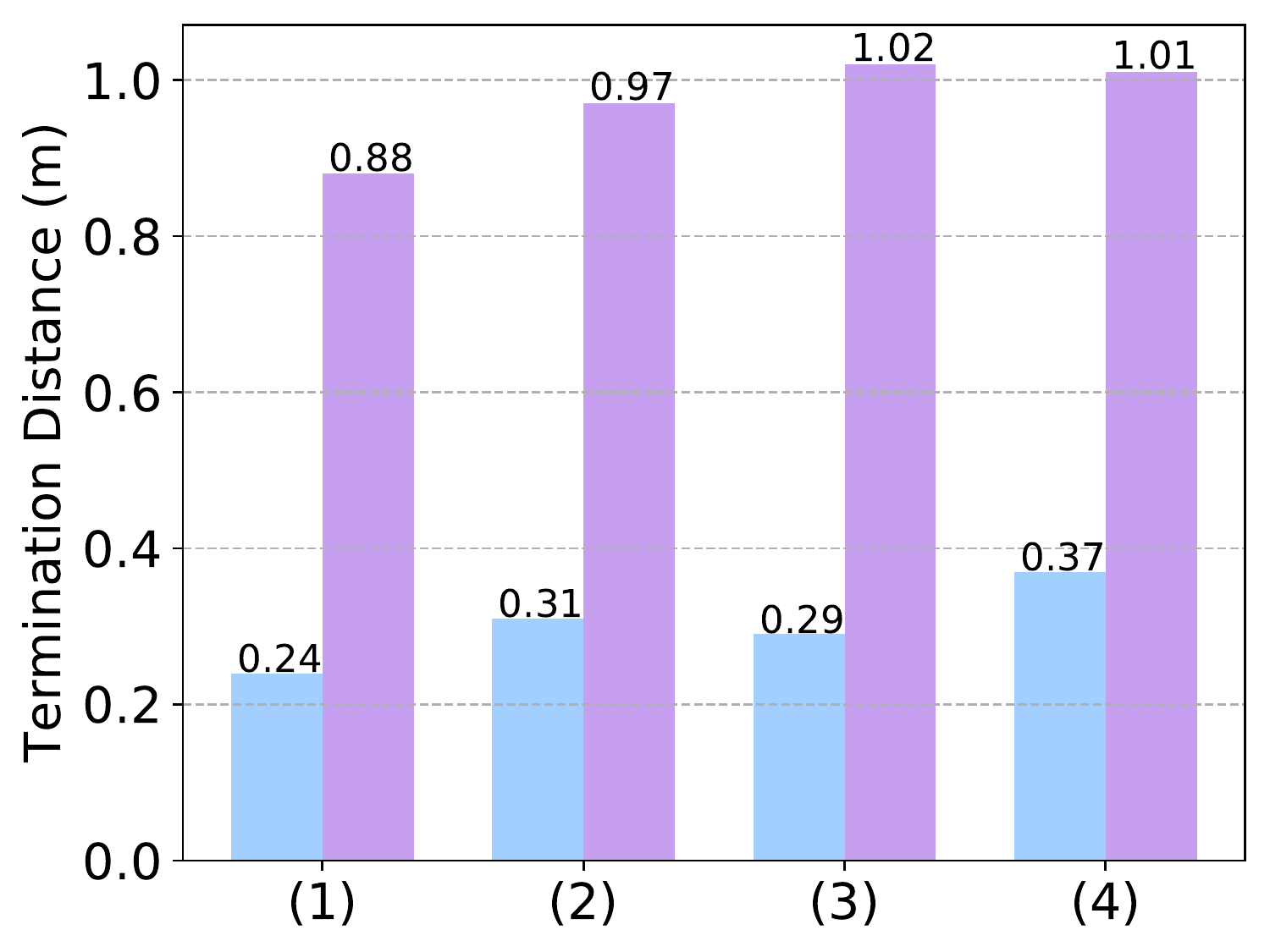}
\end{minipage}}

\caption{Transfer attack on a black-box renderer for task $T_{-10}$. Methods (1) to (4) represent PACMAN-RL+Q, NAV-GRU, NAV-Reactive, and VIS-VGG, respectively.}
\label{nav_t10}
\end{figure}

\begin{figure}[htbp]
\centering
\subfigure[Accuracy]{
\begin{minipage}[b]{0.23\linewidth}
\includegraphics[width=1\linewidth]{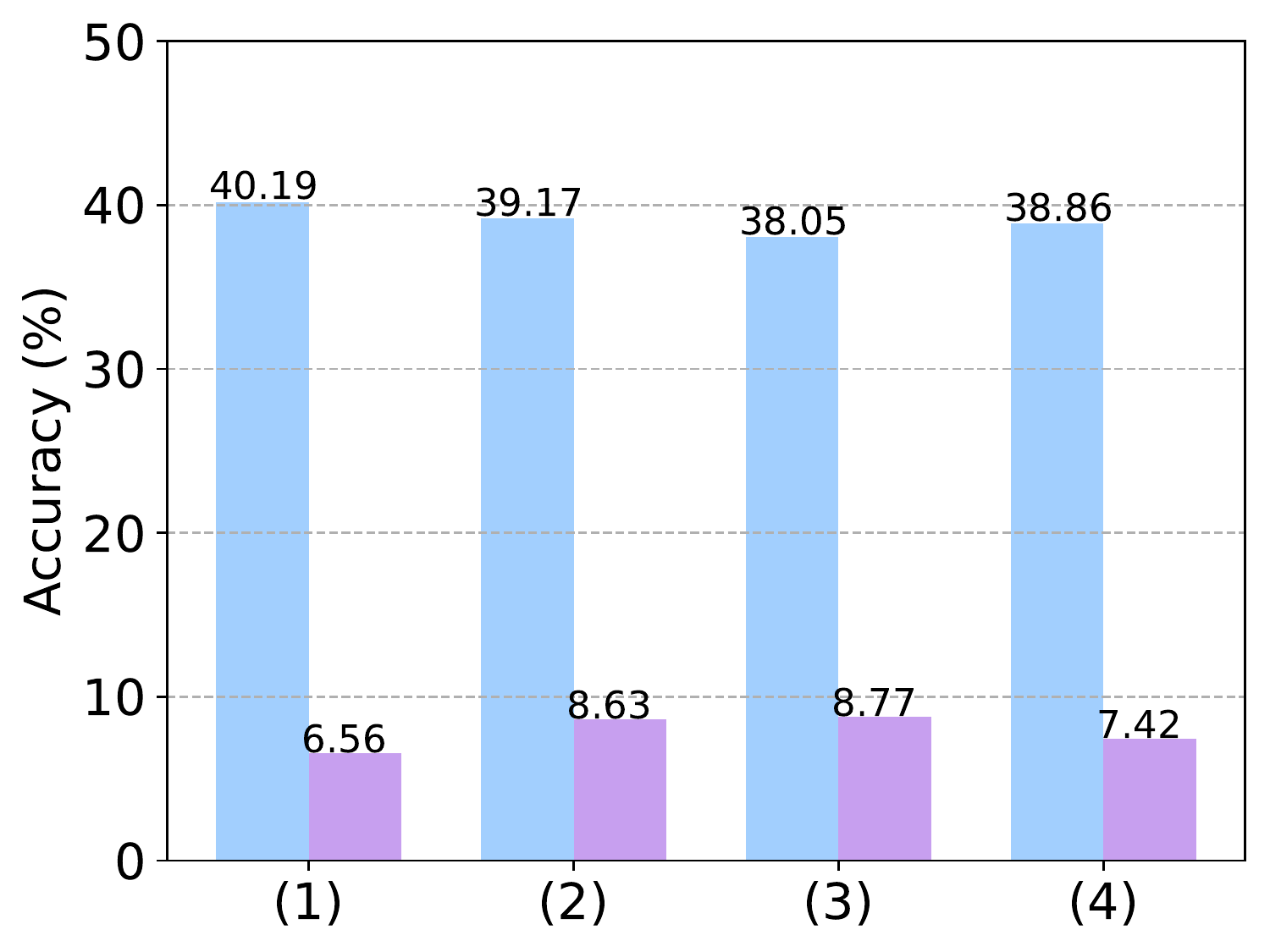}
\end{minipage}}
\subfigure[$d_T$]{
\begin{minipage}[b]{0.23\linewidth}
\includegraphics[width=1\linewidth]{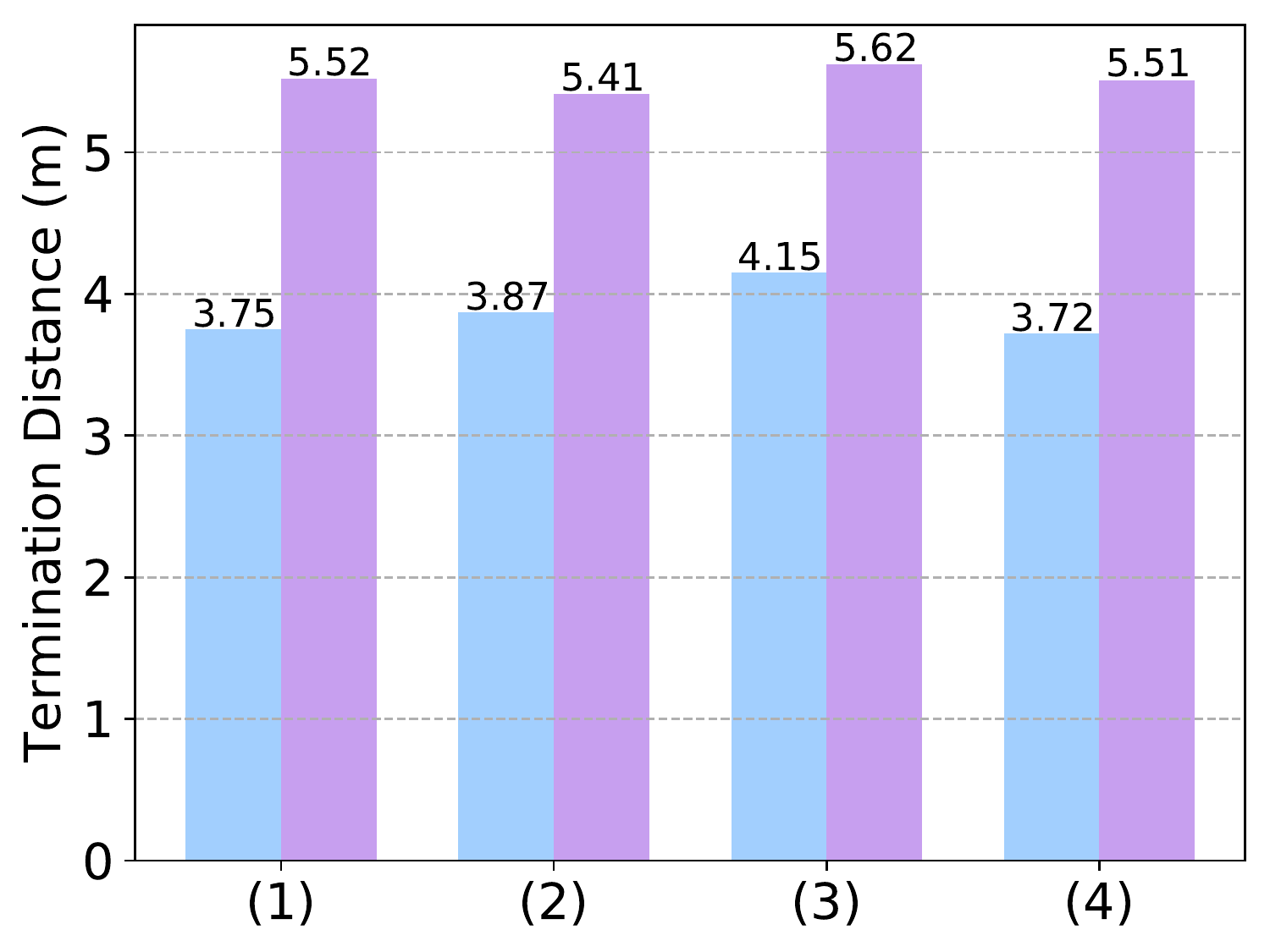}
\end{minipage}}
\subfigure[$d_{\Delta}$]{
\begin{minipage}[b]{0.23\linewidth}
\includegraphics[width=1\linewidth]{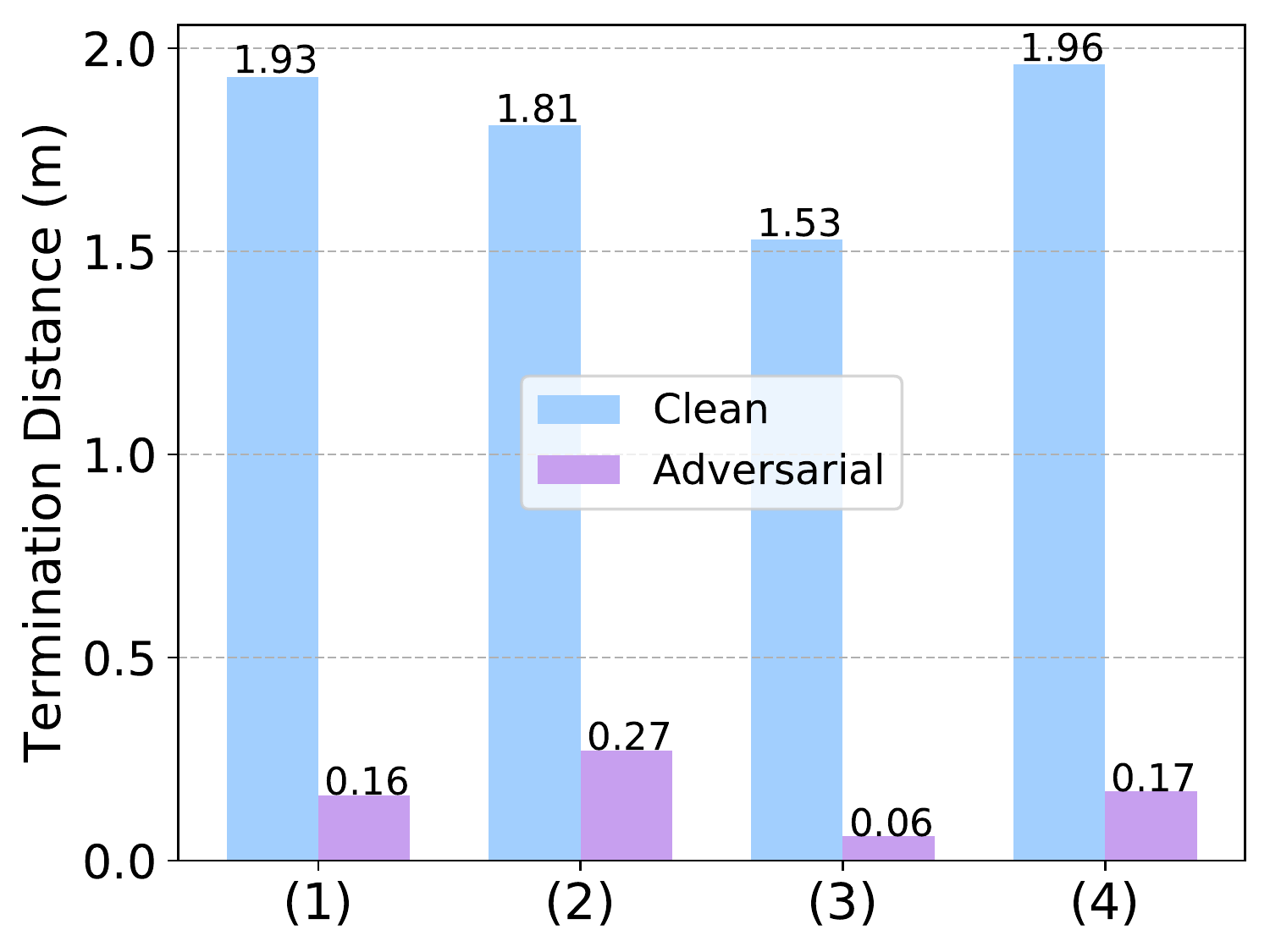}
\end{minipage}}
\subfigure[$d_{min}$]{
\begin{minipage}[b]{0.23\linewidth}
\includegraphics[width=1\linewidth]{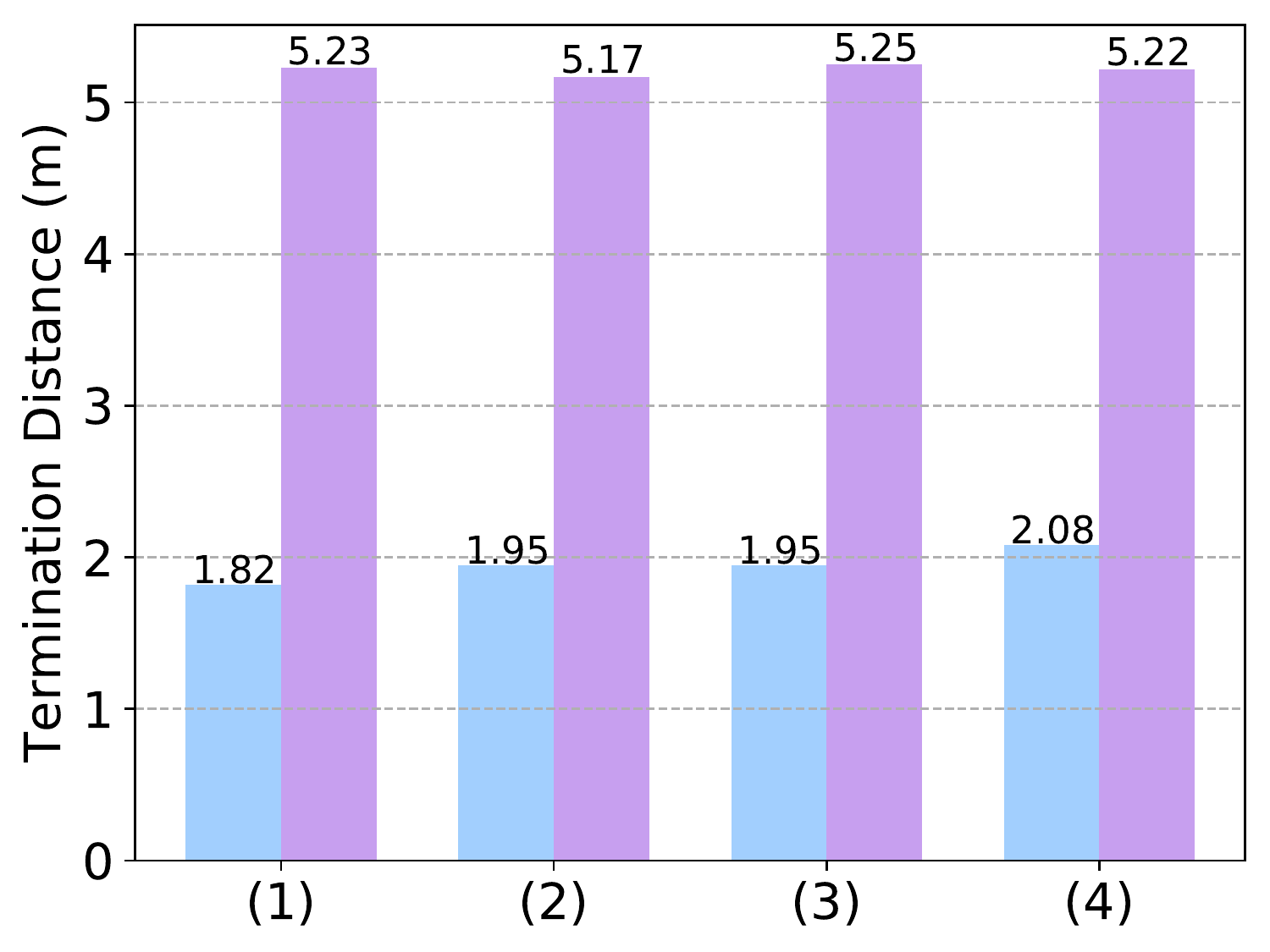}
\end{minipage}}
\caption{Transfer attack on a black-box renderer for task $T_{-50}$. Methods (1) to (4) represent PACMAN-RL+Q, NAV-GRU, NAV-Reactive, and VIS-VGG, respectively.}
\label{nav_t50}
\end{figure}

Here, we present more experimental results of transferring our generated spatiotemporal perturbations to attack a black-box renderer for EQA tasks. In addition to the results of $T_{-30}$ task discussed in Section 5.5, we further show the results of $T_{-10}$ and $T_{-50}$ tasks in Figure \ref{nav_t10} and \ref{nav_t50}, respectively. Again, we observe that our attacks transfer to the black-box renderer.

\subsection{Sample Adversarial Attacks for Question Answering and Visual Recognition}

\begin{figure}[htbp]
\subfigure[Clean scene]{
\begin{minipage}[b]{0.48\linewidth}
\includegraphics[width=1\linewidth]{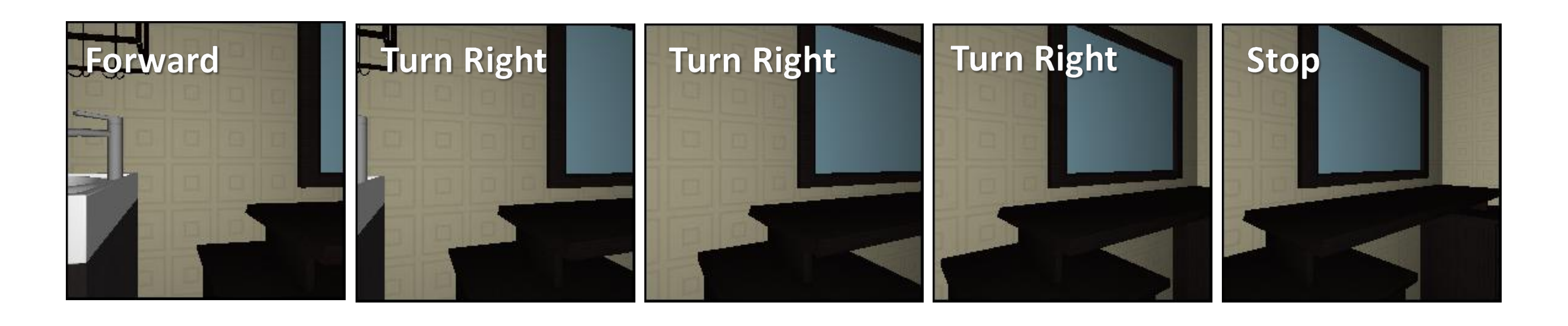}
\end{minipage}}
\subfigure[Adversarial scene]{
\begin{minipage}[b]{0.48\linewidth}
\includegraphics[width=1\linewidth]{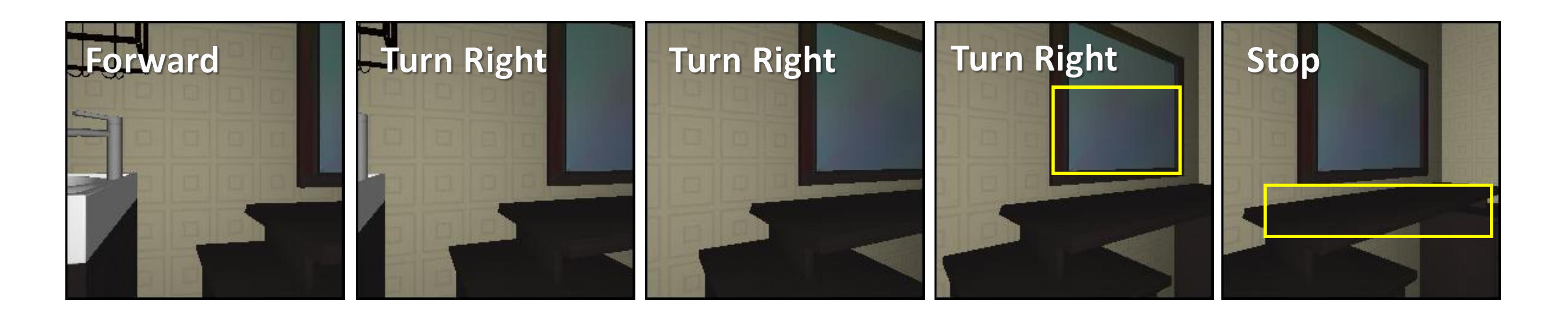}
\end{minipage}}

\caption{Given the question ``\emph{What color is the mirror?}'', we show the last 5 views of the agent for EQA in
the same scene with and without adversarial perturbations. The contextual objects perturbed including table and mirror. The ground truth prediction is ``white''. The agent gives the wrong answer ``black'' to the question.} 
\label{Mirror}
\end{figure}

\begin{figure}[htbp]
\centering
\subfigure[Clean scene]{
\begin{minipage}[b]{0.48\linewidth}

\includegraphics[width=1\linewidth]{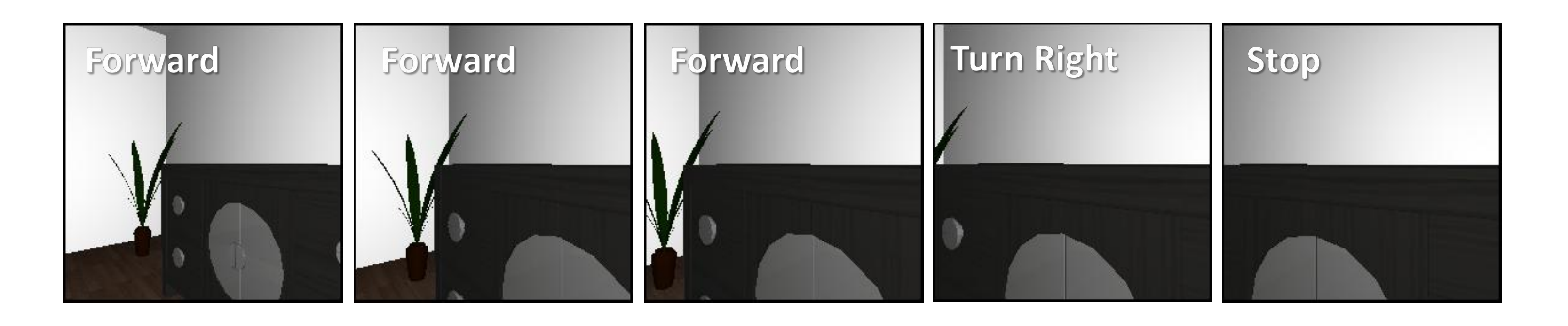}
\end{minipage}}
\subfigure[Adversarial scene]{
\begin{minipage}[b]{0.48\linewidth}

\includegraphics[width=1\linewidth]{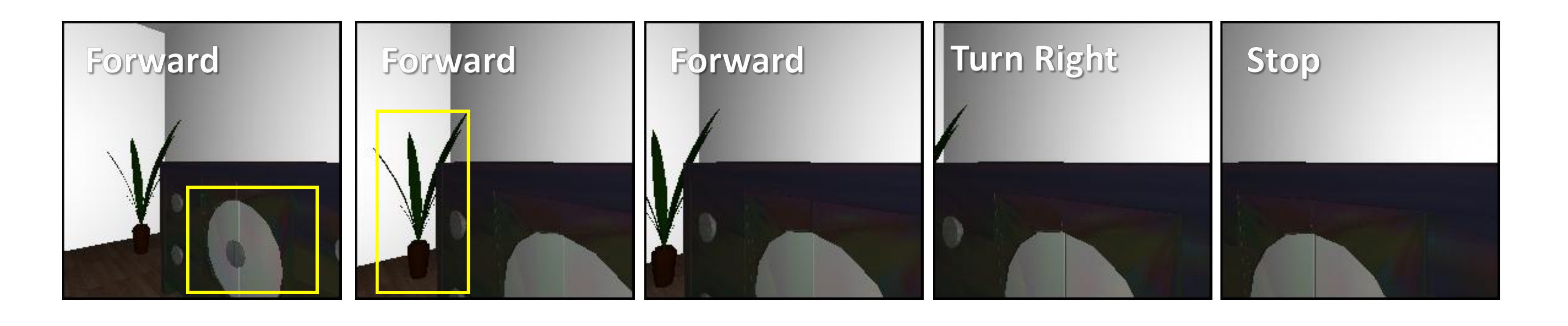}
\end{minipage}}

\caption{Given the question ``\emph{What color is the shoes cabinet?}'', we show the last 5 views of the agent for EQA in
the same scene with and without adversarial perturbations. The contextual objects perturbed including cabinet and plant. The ground truth prediction is ``brown''. The agent gives the wrong answer ``yellow'' to the question.}
\label{Cabinet}
\end{figure}

\begin{figure}[htbp]
\centering
\subfigure[Clean scene]{
\begin{minipage}[b]{0.48\linewidth}

\includegraphics[width=1\linewidth]{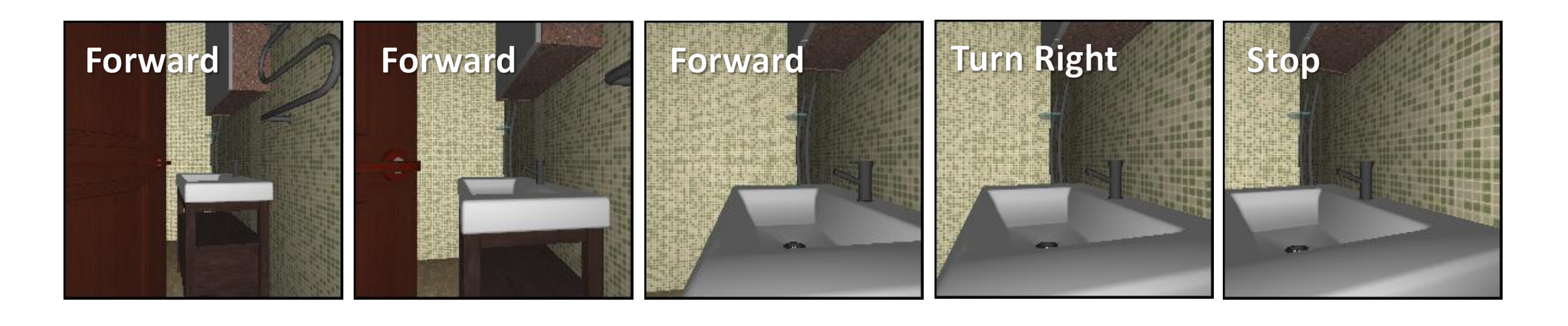}
\end{minipage}}
\subfigure[Adversarial scene]{
\begin{minipage}[b]{0.48\linewidth}

\includegraphics[width=1\linewidth]{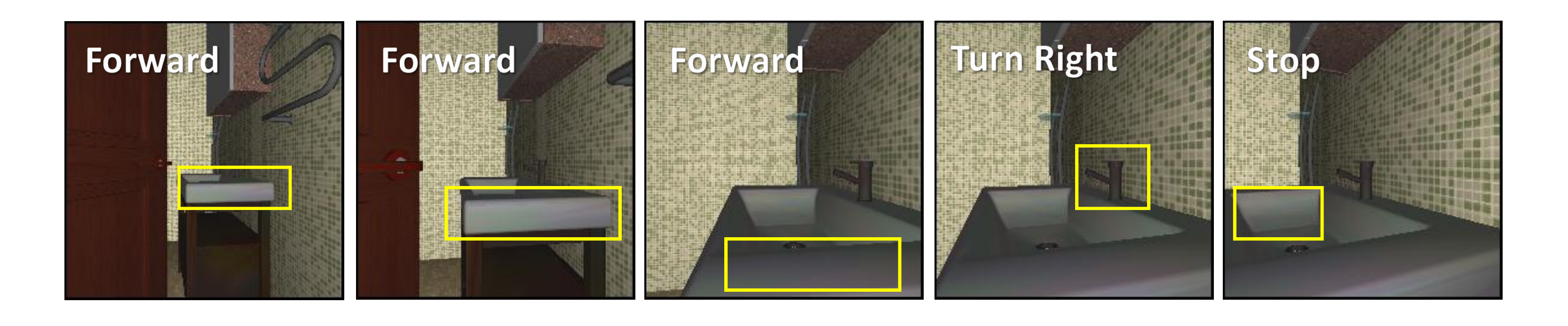}
\end{minipage}}

\caption{Given the question ``\emph{What room is the sink located in?}'', we show the last 5 views of the agent for EQA in
the same scene with and without adversarial perturbations. The contextual objects perturbed including sink and water tap. The ground truth prediction is ``bathroom''. The agent gives the wrong answer ``kitchen'' to the question.}
\label{Sink}
\end{figure}

\begin{figure}[htbp]
\centering
\subfigure[Clean scene]{
\begin{minipage}[b]{0.48\linewidth}

\includegraphics[width=1\linewidth]{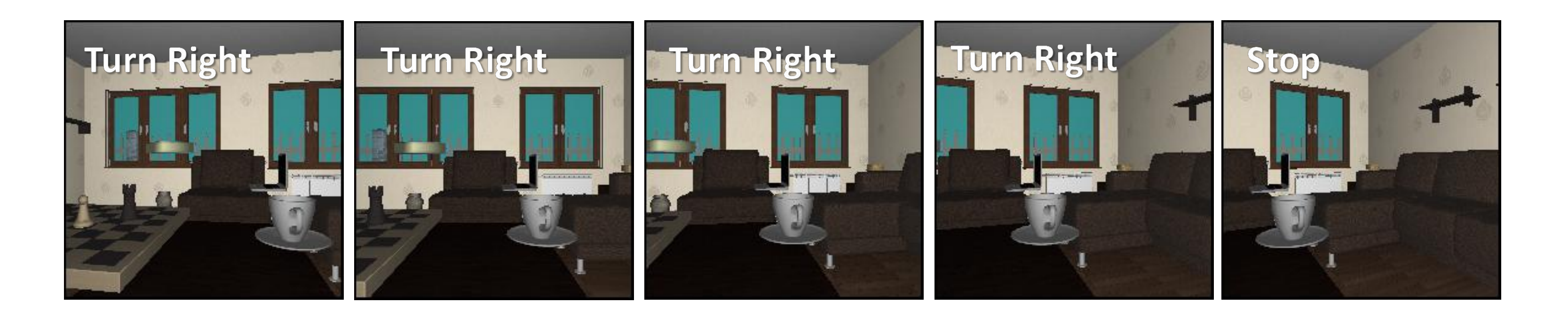}
\end{minipage}}
\subfigure[Adversarial scene]{
\begin{minipage}[b]{0.48\linewidth}

\includegraphics[width=1\linewidth]{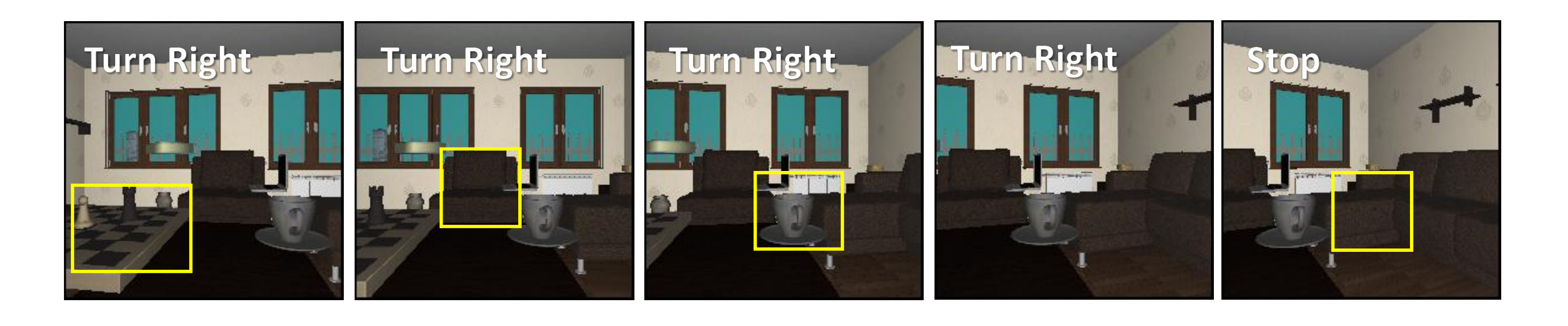}
\end{minipage}}
\caption{Given the question ``\emph{What room is the cup located in?}'', we show the last 5 views of the agent for EQA in
the same scene with and without adversarial perturbations. The contextual objects perturbed including chessboard, cup, and sofa. The ground truth prediction is ``living room''. The agent gives the wrong answer ``bedroom'' to the question.}
\label{Cup}
\end{figure}

\begin{figure}[htb]
\vspace{-0.05in}
\centering
\subfigure[Clean Scene]{
\begin{minipage}[b]{0.49\linewidth}
\includegraphics[width=1\linewidth]{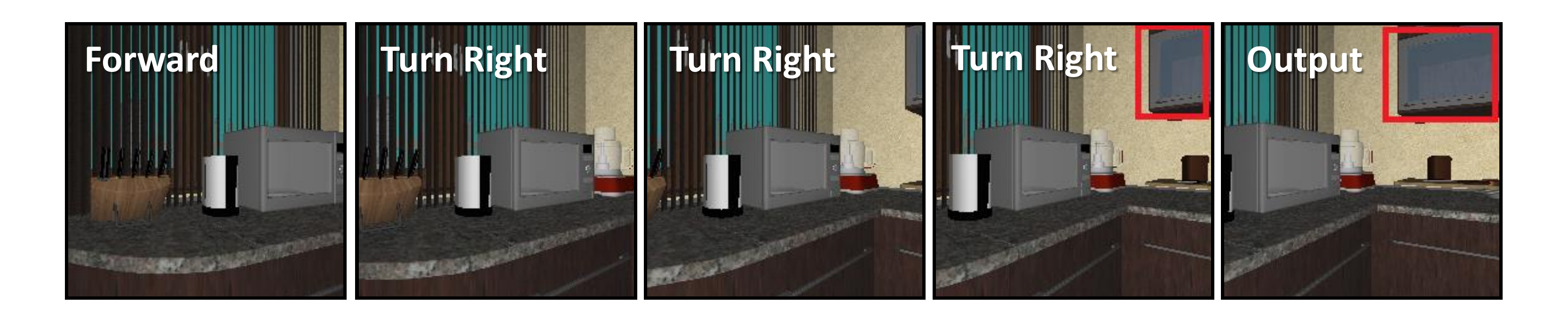}
\end{minipage}}
\subfigure[Adversarial Scene]{
\begin{minipage}[b]{0.49\linewidth}
\includegraphics[width=1\linewidth]{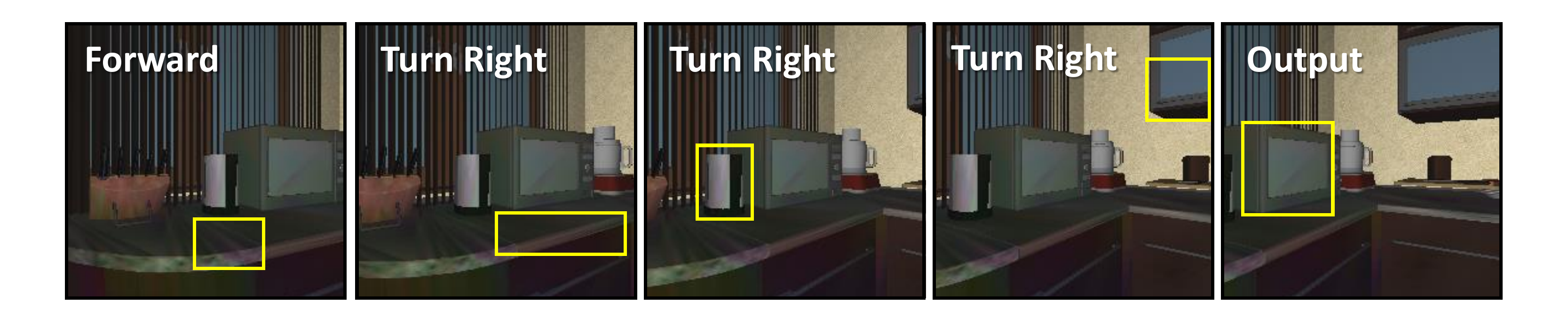}
\end{minipage}}
\vspace{-0.2in}
	\caption{The last 5 views of the agent for EVR in the same scene with and without adversarial perturbations. The contextual objects perturbed are table, kettle, microwave, and cabinet. After the adversarial attack, the agent fails to recognize the cabinet in subfigure (b). Red boxes indicate the bounding box for object detection and yellow boxes show the adversarially perturbed texture regions.}
	\label{fig:perceptual_evr}
\vspace{-0.15in}
\end{figure}

\begin{figure}[htbp]
\centering
\subfigure[Clean scene]{
\begin{minipage}[b]{0.48\linewidth}

\includegraphics[width=1\linewidth]{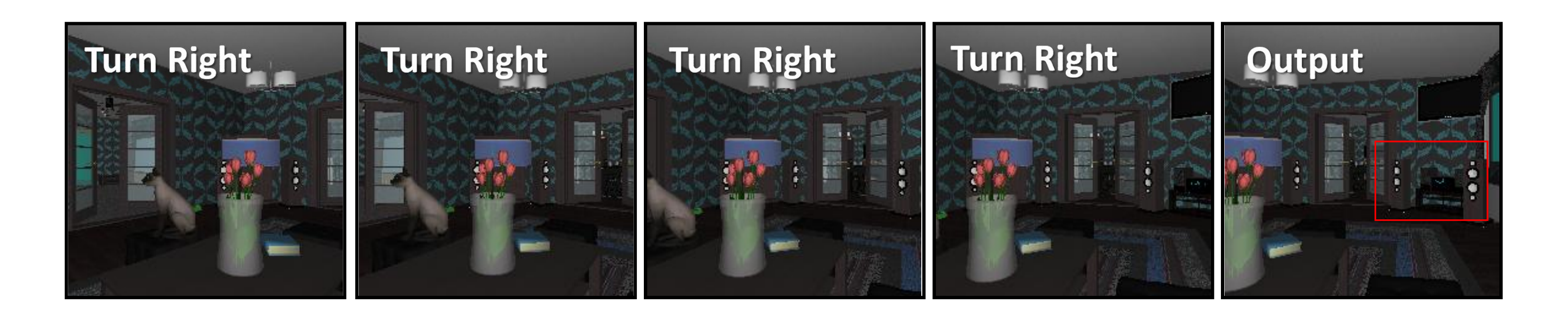}
\end{minipage}}
\subfigure[Adversarial scene]{
\begin{minipage}[b]{0.48\linewidth}

\includegraphics[width=1\linewidth]{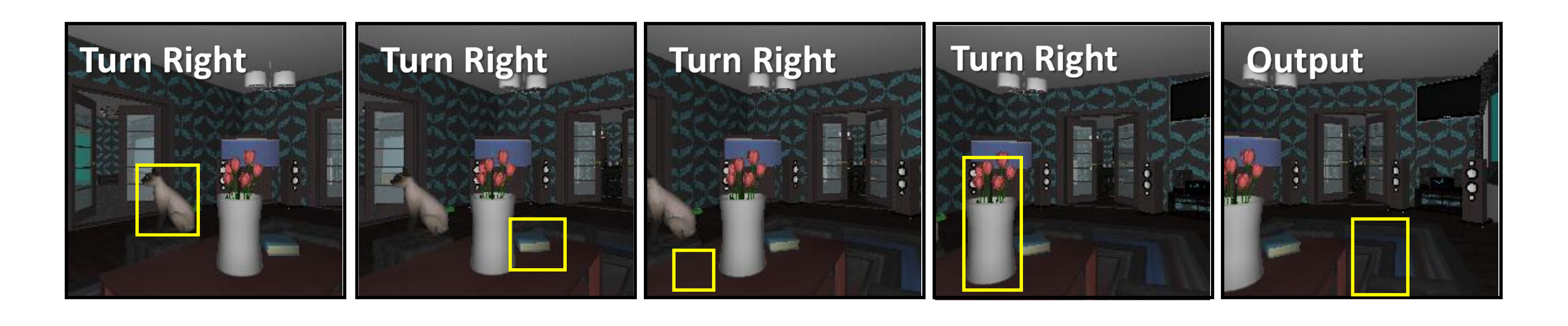}
\end{minipage}}

\caption{The last 5 views of the agent for EVR in the same scene with and without adversarial perturbations. The contextual objects perturbed are dog, book, desk, vase, and carpet. After the adversarial attack, the agent fails to recognize the stereoset in subfigure (b). Red boxes indicate the bounding box for object detection and yellow boxes show the adversarially perturbed texture regions.}
\label{evr1}
\end{figure}

\begin{figure}[htbp]
\centering
\subfigure[Clean scene]{
\begin{minipage}[b]{0.48\linewidth}

\includegraphics[width=1\linewidth]{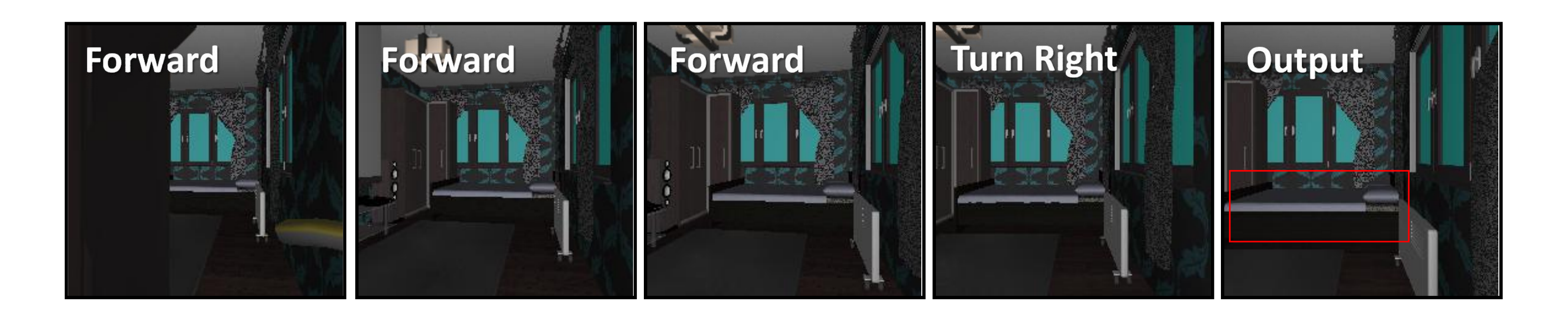}
\end{minipage}}
\subfigure[Adversarial scene]{
\begin{minipage}[b]{0.48\linewidth}

\includegraphics[width=1\linewidth]{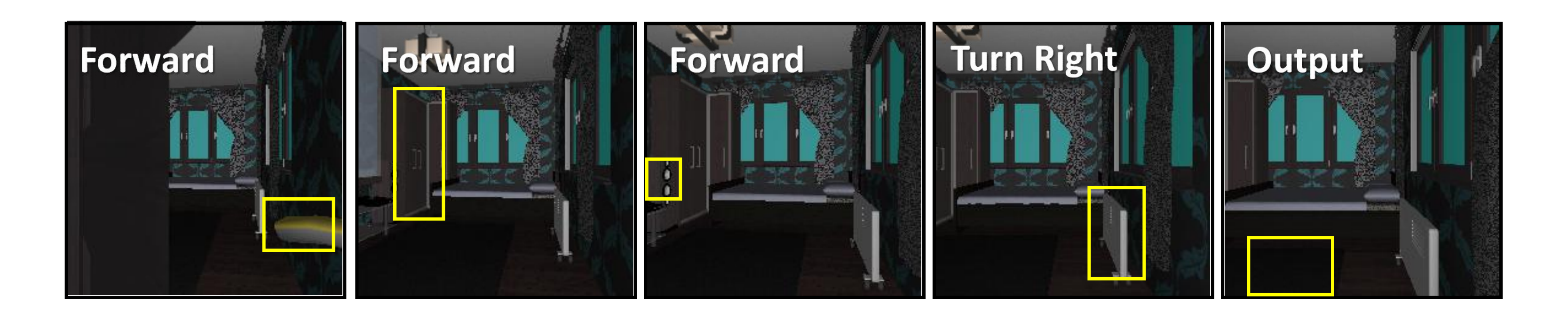}
\end{minipage}}

\caption{The last 5 views of the agent for EVR in the same scene with and without adversarial perturbations. The contextual objects perturbed are sofa, cabinet, stereo set, heating, and carpet. After the adversarial attack, the agent fails to recognize the bed in subfigure (b). Red boxes indicate the bounding box for object detection and yellow boxes show the adversarially perturbed texture regions.}
\label{evr2}
\end{figure}

In this section, we show more examples of adversarial scenes generated using our attack framework. Figures \ref{Mirror}, \ref{Cabinet}, \ref{Sink}, and \ref{Cup} illustrate some examples of our adversarial attacks for question answering. All of these questions are answered correctly by agents in clean scenes, but wrongly in corresponding adversarial scenes. Examples for visual recognition are shown in Figures \ref{fig:perceptual_evr}, \ref{evr1} and \ref{evr2}. All of these objects are classified correctly by agents in clean scenes, but wrongly in corresponding adversarial scenes.

\subsection{Sample Attention Maps of the Agent for Question Answering}

\begin{figure}[htbp]
\centering
\subfigure[Clean scene]{
\begin{minipage}[b]{0.48\linewidth}

\includegraphics[width=1\linewidth]{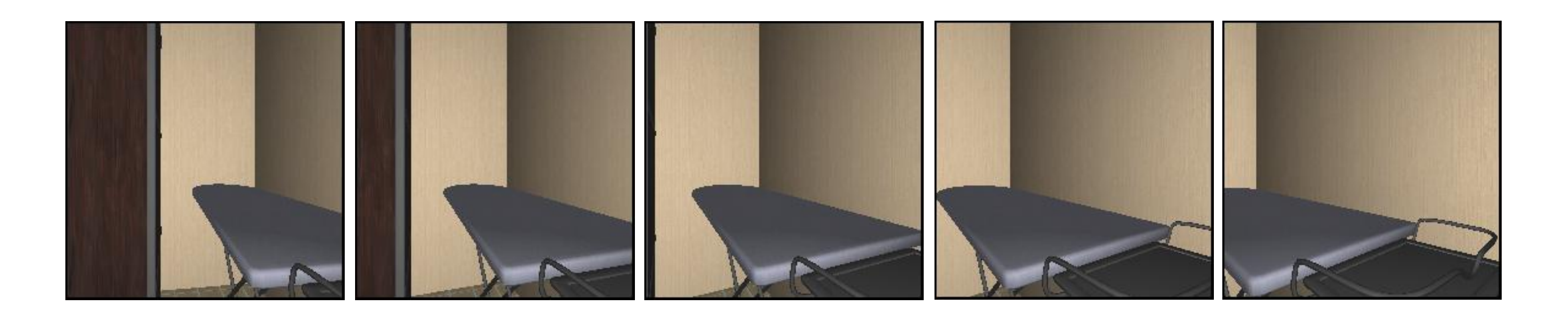}
\end{minipage}}
\subfigure[Attention in clean scene]{
\begin{minipage}[b]{0.48\linewidth}

\includegraphics[width=1\linewidth]{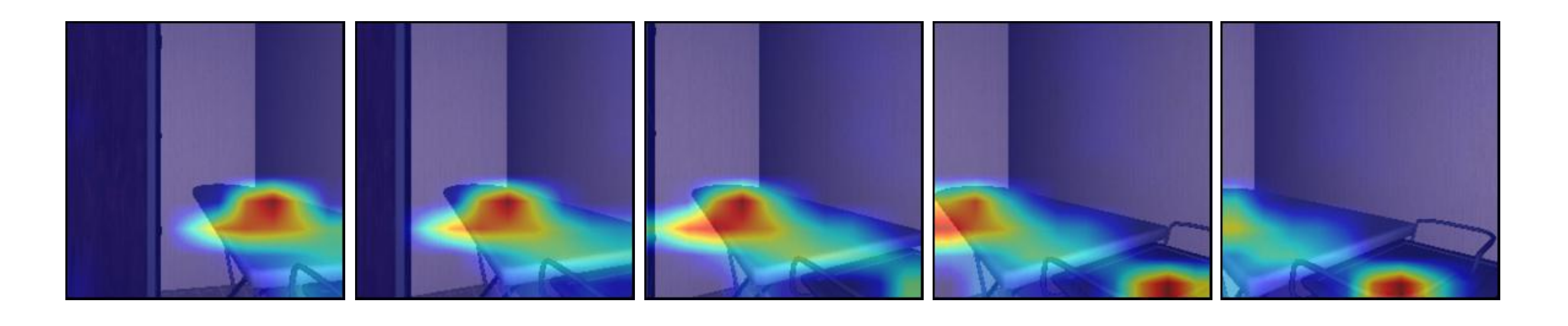}
\end{minipage}}

\subfigure[Adversarial scene]{
\begin{minipage}[b]{0.48\linewidth}

\includegraphics[width=1\linewidth]{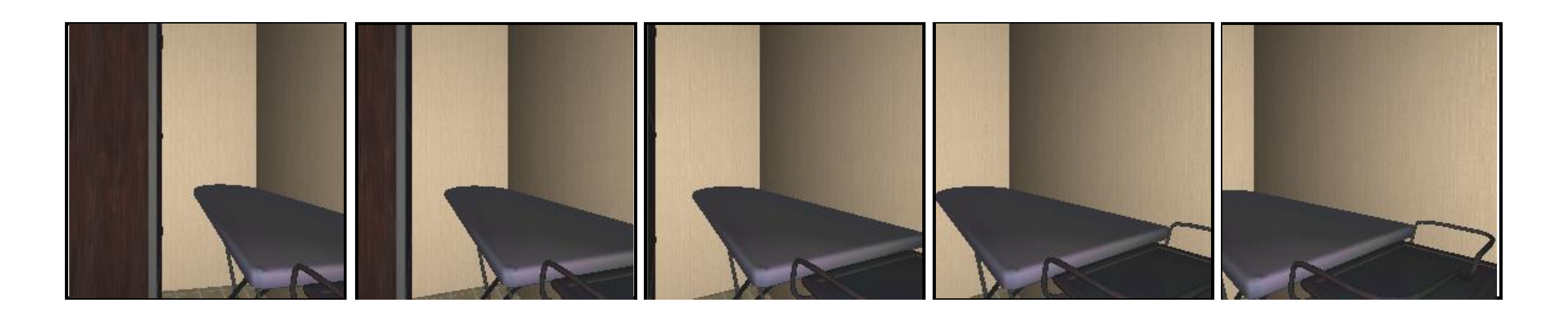}
\end{minipage}}
\subfigure[Attention in adversarial scene]{
\begin{minipage}[b]{0.48\linewidth}

\includegraphics[width=1\linewidth]{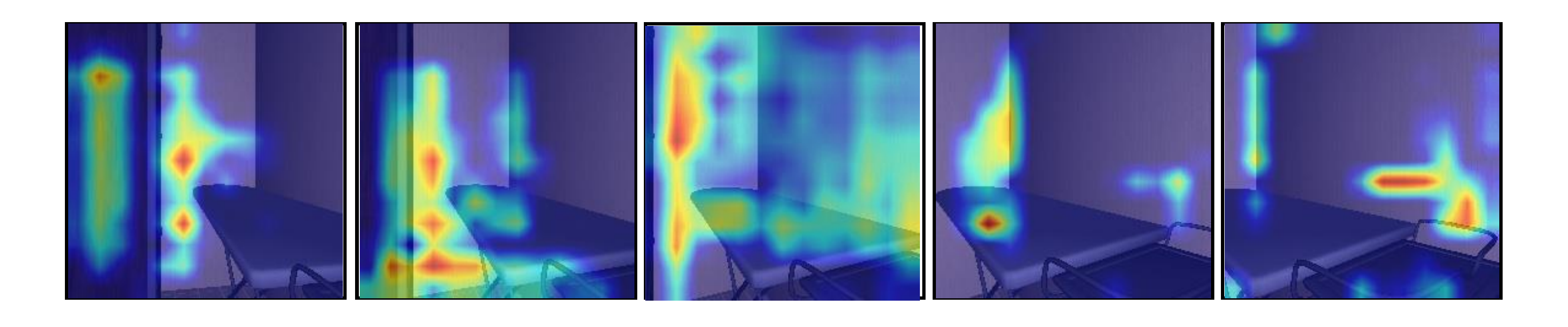}
\end{minipage}}

\caption{Egocentric views and corresponding attention maps when the agent answers the question, ``\emph{What color is the ironing board?}''. The agent mainly focuses on the target object when predicting its color in the clean scene (subfigure (a) and (b)). The adversarial scene and corresponding attention maps are shown in subfigure (c) and (d). The ground truth prediction is 'white'. The agent gives the wrong answer 'brown' to the question.}
\label{IronBoard_cam}
\end{figure}
\begin{figure}[htbp]
\centering
\subfigure[Clean scene]{
\begin{minipage}[b]{0.48\linewidth}

\includegraphics[width=1\linewidth]{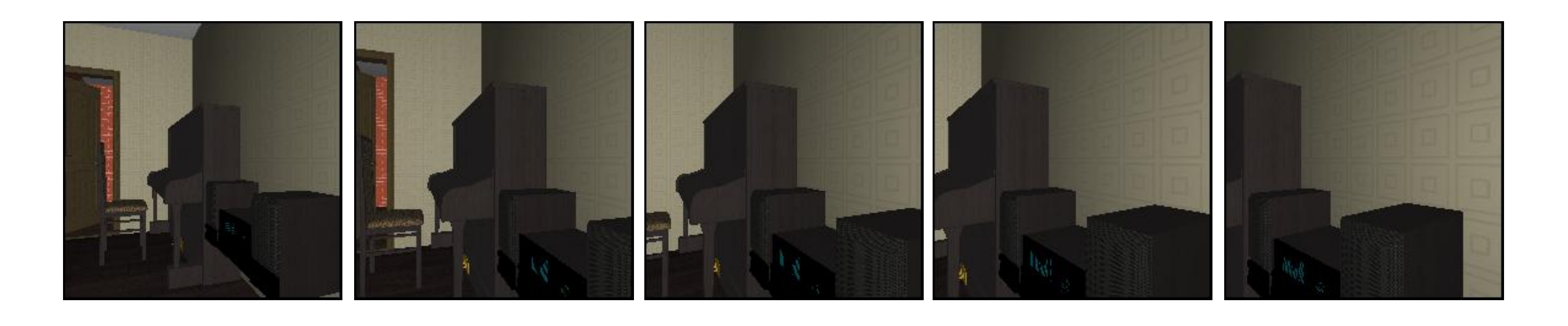}
\end{minipage}}
\subfigure[Attention in clean scene]{
\begin{minipage}[b]{0.48\linewidth}

\includegraphics[width=1\linewidth]{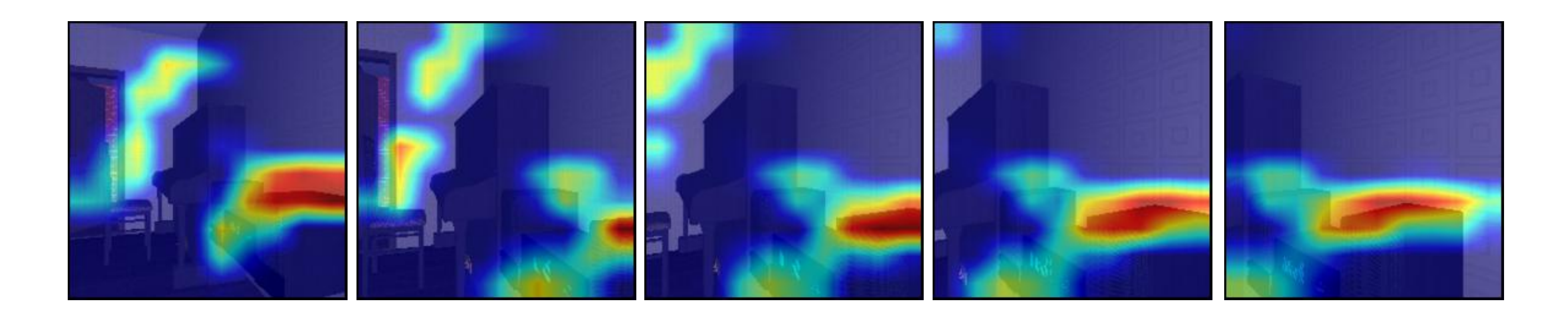}
\end{minipage}}

\subfigure[Adversarial scene]{
\begin{minipage}[b]{0.48\linewidth}

\includegraphics[width=1\linewidth]{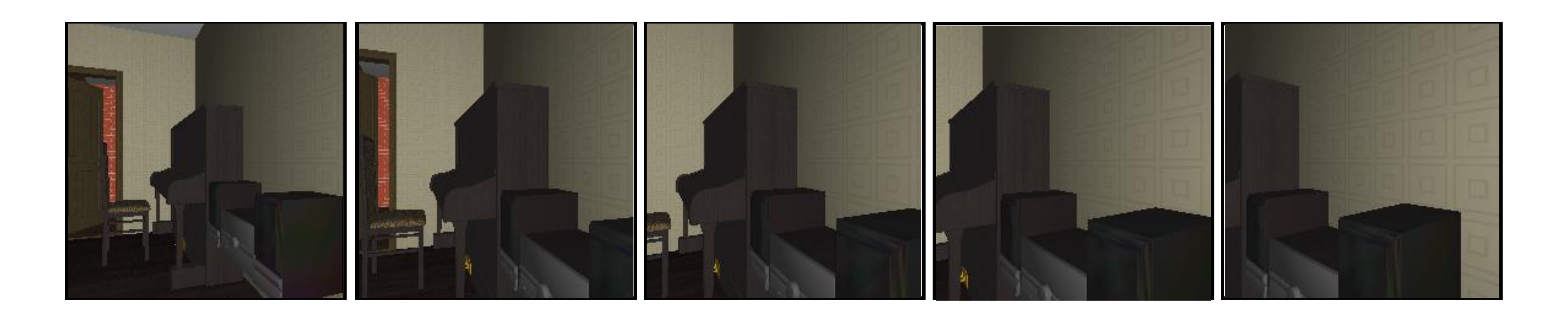}
\end{minipage}}
\subfigure[Attention in adversarial scene]{
\begin{minipage}[b]{0.48\linewidth}

\includegraphics[width=1\linewidth]{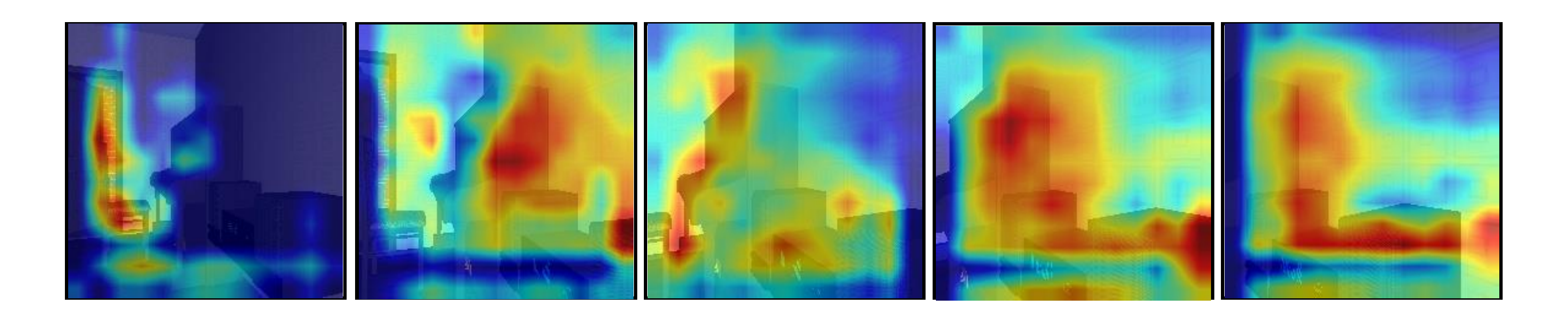}
\end{minipage}}

\caption{Egocentric views and corresponding attention maps when the agent answers the question, ``\emph{What color is the stereo set?}''. The agent mainly focuses on the target object when predicting its color in the clean scene (subfigure (a) and (b)). The adversarial scene and corresponding attention maps are shown in subfigure (c) and (d). The ground truth prediction is 'black'. The agent gives the wrong answer 'white' to the question.}
\label{StereoSet_cam}
\end{figure}
\begin{figure}[htbp]
\centering
\subfigure[Clean scene]{
\begin{minipage}[b]{0.48\linewidth}

\includegraphics[width=1\linewidth]{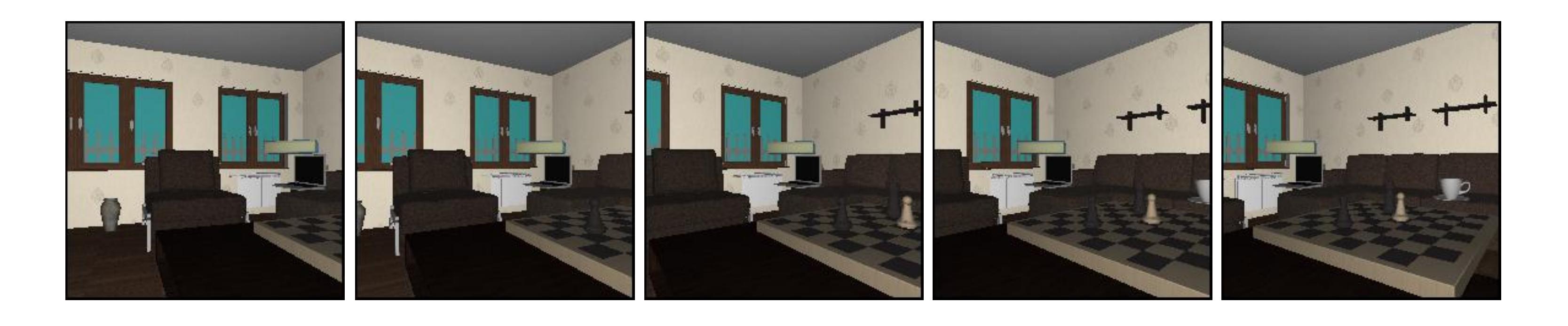}
\end{minipage}}
\subfigure[Attention in clean scene]{
\begin{minipage}[b]{0.48\linewidth}

\includegraphics[width=1\linewidth]{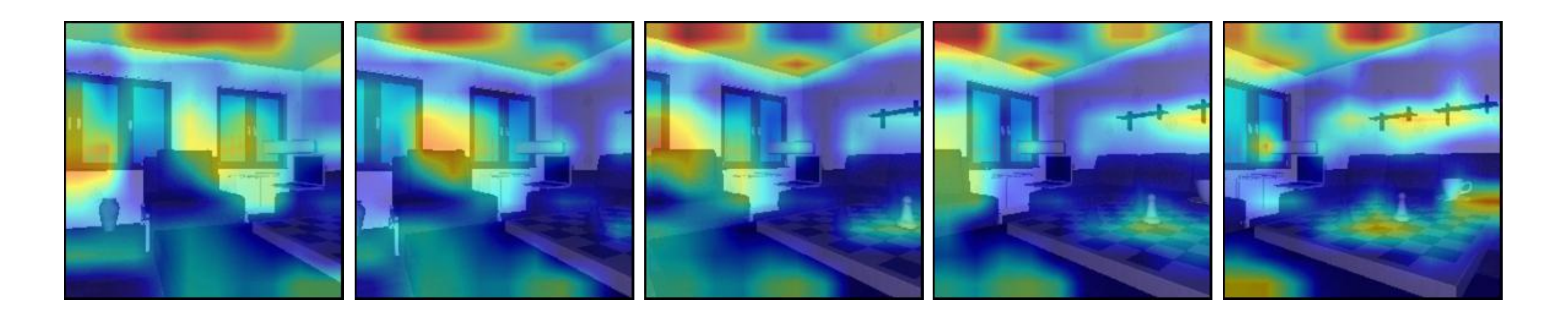}
\end{minipage}}
\subfigure[Adversarial scene]{
\begin{minipage}[b]{0.48\linewidth}

\includegraphics[width=1\linewidth]{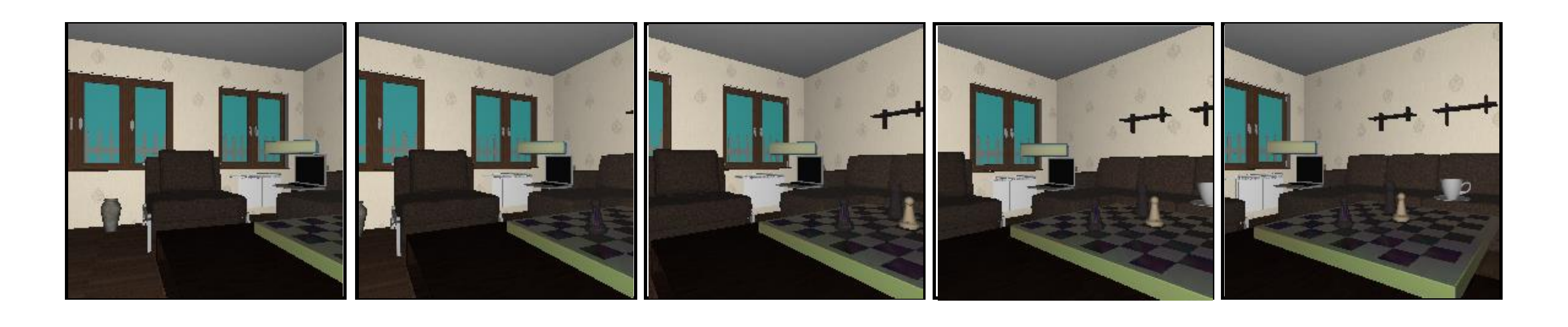}
\end{minipage}}
\subfigure[Attention in adversarial scene]{
\begin{minipage}[b]{0.48\linewidth}

\includegraphics[width=1\linewidth]{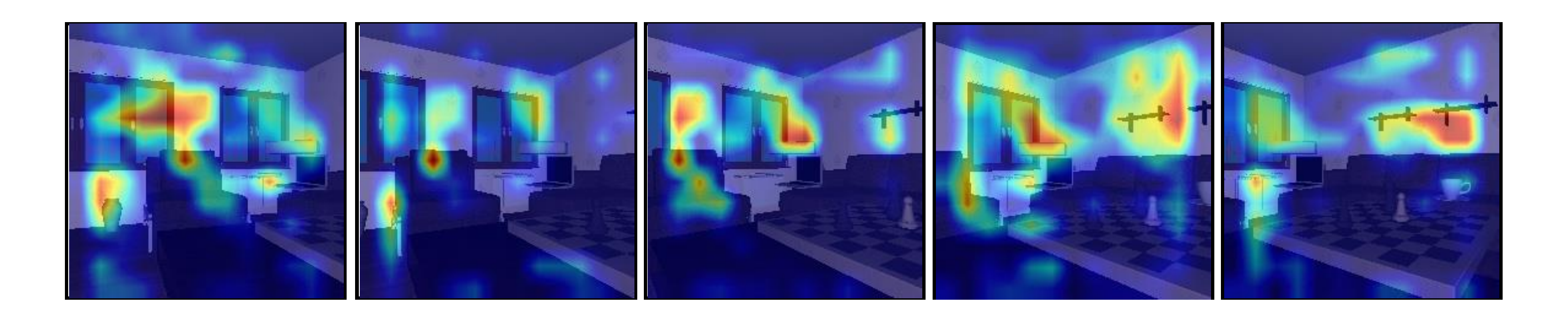}
\end{minipage}}
\caption{Egocentric views and corresponding attention maps when the agent answers the question, ``\emph{What room is the chessboard located in?}''. The agent uses clues from contextual objects to answer locational and compositional questions in the clean scene (subfigure (a) and (b)). The adversarial scene and corresponding attention maps are shown in subfigure (c) and (d). The ground truth prediction is 'living room'. The agent gives the wrong answer 'bathroom' to the question.}
\label{ChessBoard_cam}
\end{figure}
\begin{figure}[htbp]
\centering
\subfigure[Clean scene]{
\begin{minipage}[b]{0.48\linewidth}

\includegraphics[width=1\linewidth]{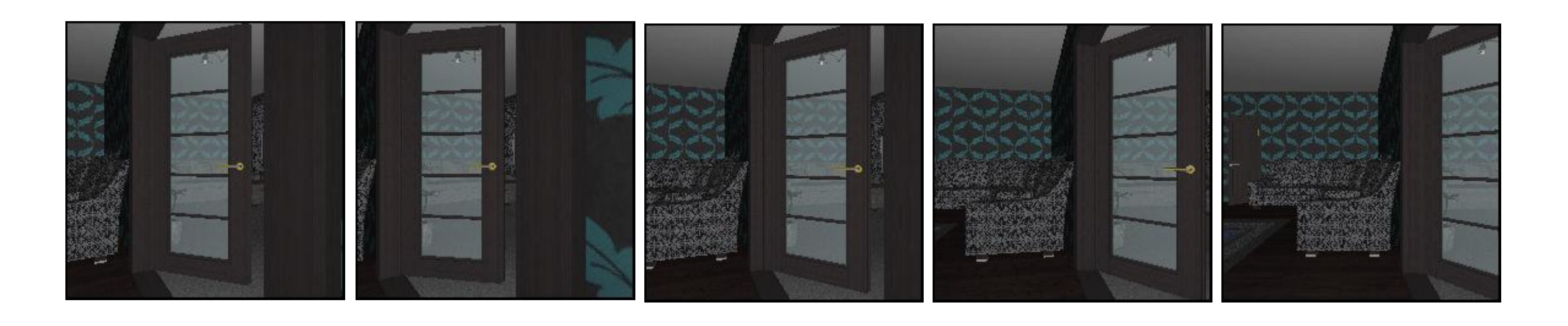}
\end{minipage}}
\subfigure[Attention in clean scene]{
\begin{minipage}[b]{0.48\linewidth}

\includegraphics[width=1\linewidth]{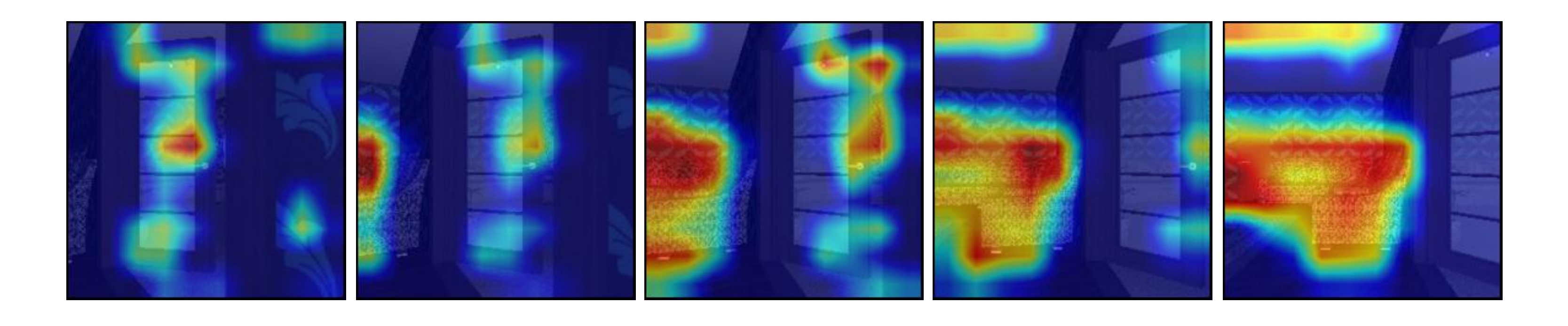}
\end{minipage}}

\subfigure[Adversarial scene]{
\begin{minipage}[b]{0.48\linewidth}

\includegraphics[width=1\linewidth]{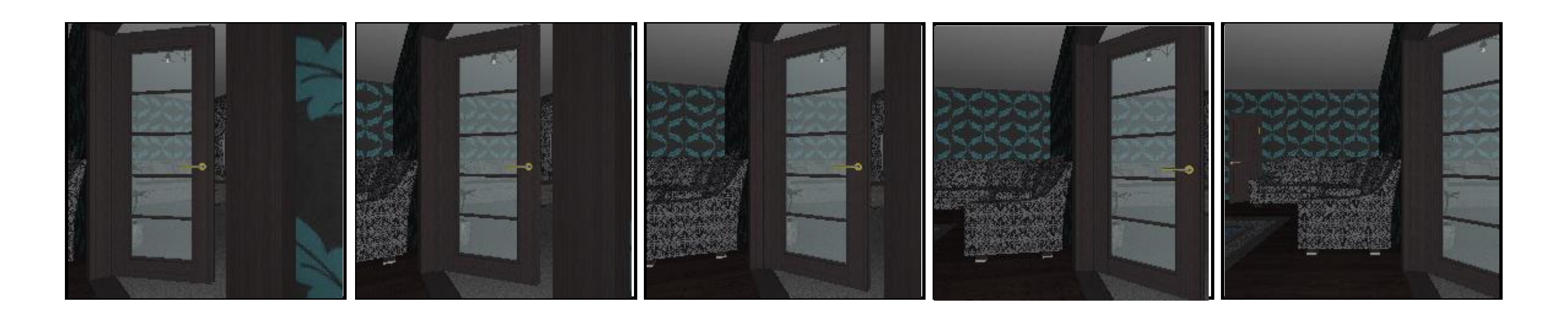}
\end{minipage}}
\subfigure[Attention in adversarial scene]{
\begin{minipage}[b]{0.48\linewidth}

\includegraphics[width=1\linewidth]{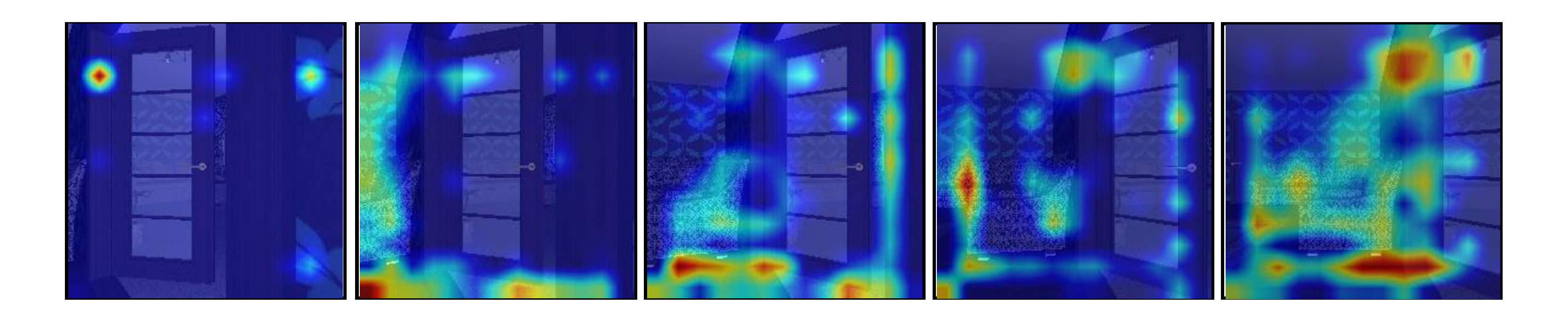}
\end{minipage}}

\caption{Egocentric views and corresponding attention maps when the agent answers the question, ``\emph{What room is the toy located in?}''. The agent uses clues from contextual objects to answer locational and compositional questions  in the clean scene (subfigure (a) and (b)). The adversarial scene and corresponding attention maps are shown in subfigure (c) and (d). The ground truth prediction is 'living room'. The agent gives the wrong answer 'kitchen' to the question.}
\label{toy_cam}
\end{figure}

As shown in Figures \ref{IronBoard_cam}, \ref{StereoSet_cam}, \ref{ChessBoard_cam}, and \ref{toy_cam}, we provide more visualization of the egocentric views and corresponding attention maps when agents answer questions. We can observe that the agents use clues from contextual objects to answer locational and compositional questions while mainly focus on target objects when predicting their colors.

\begin{figure*}[htbp]
	\centering
\subfigure[Clean scene]{
\includegraphics[width=0.95\linewidth]{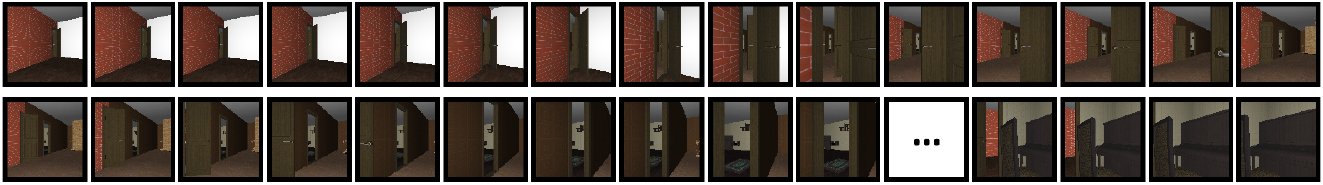}
}

\subfigure[Adversarial scene]{
\includegraphics[width=0.95\linewidth]{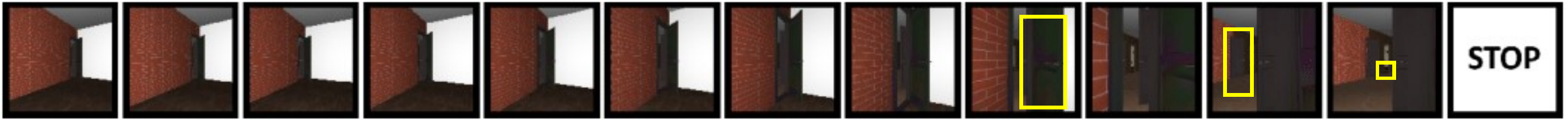}
}
\caption{Egocentric views of the agent in the same scene with and without adversarial perturbations. As shown in subfigure (b), we perturb the textures of two doors and the rug. The agent stops at the 5-$th$ \emph{planner} step.}
\label{nav_attack_0}
\end{figure*}

\begin{figure*}[htbp]
	\centering
\subfigure[Clean scene]{
\includegraphics[width=0.95\linewidth]{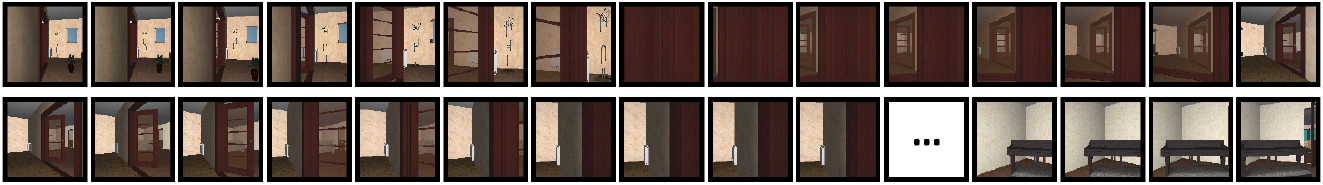}
}

\subfigure[Adversarial scene]{
\includegraphics[width=0.95\linewidth]{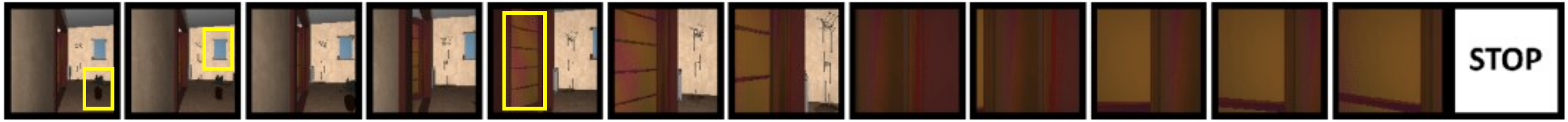}
}
\caption{Egocentric views of the agent in the same scene with and without adversarial perturbations. As shown in subfigure (b), we perturb the textures of the plant, the door, and the window. The agent stops at the 4-$th$ \emph{planner} step.}
\label{nav_attack_1}
\end{figure*}

\begin{figure*}[htbp]
	\centering
\subfigure[Clean scene]{
\includegraphics[width=0.95\linewidth]{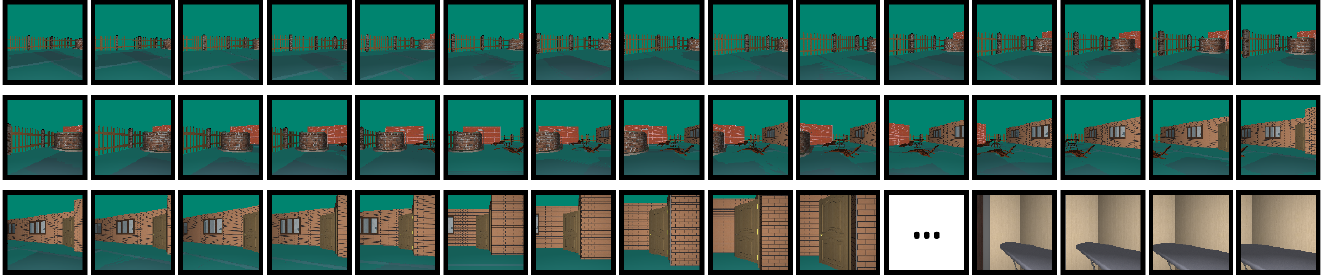}
}

\subfigure[Adversarial scene]{
\includegraphics[width=0.95\linewidth]{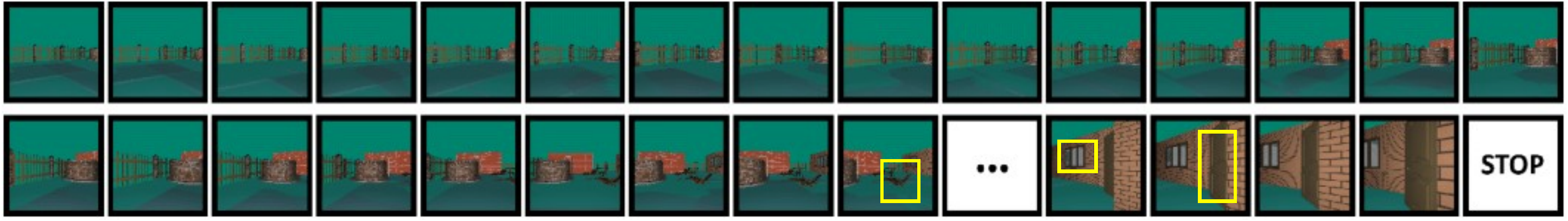}
}
\caption{Egocentric views of the agent in the same scene with and without adversarial perturbations. As shown in subfigure (b), we perturb the textures of the bench, the door, and the window. The agent stops at the 12-$th$ \emph{planner} step.}
\label{nav_attack_2}
\end{figure*}

\subsection{Sample Adversarial Attacks for Navigation}

In this section, we show more examples of adversarial scenes for navigation in Figures \ref{nav_attack_0}, \ref{nav_attack_1}, and \ref{nav_attack_2}. Agents navigate correctly to the end in all of the clean scenes, but stop ahead of time in corresponding adversarial scenes.

\clearpage
%
%
\bibliographystyle{splncs04}
\bibliography{egbib}

\end{document}